\documentclass[10pt,journal,compsoc]{IEEEtran}
\usepackage{amsmath,amsfonts}
\usepackage{algorithmic}
\usepackage{algorithm}
\usepackage{array}
\usepackage[caption=false,font=normalsize,labelfont=sf,textfont=sf]{subfig}
\usepackage{textcomp}
\usepackage{stfloats}
\usepackage{url}
\usepackage{verbatim}
\usepackage{graphicx}
\usepackage{cite}

\usepackage{booktabs}
\usepackage{multirow}
\usepackage{xcolor}
\usepackage{balance}
\usepackage{tabularx}
\usepackage{makecell}

\usepackage{marvosym}
\usepackage{capt-of}

\usepackage[colorlinks=true,
            citecolor=green]{hyperref}

\usepackage[table]{xcolor}
\definecolor{DarkGray}{RGB}{235,235,235}
\newcolumntype{Y}{>{\centering\arraybackslash}X}
\newcommand{\SAPC}{\mbox{SA,\,PC}}
\newcommand{\graycell}[1]{\textcolor{black}{#1}}

\newcommand{\grayrowA}[7]{\graycell{#1} & \graycell{#2} & \graycell{#3} & \graycell{#4} & \graycell{#5} & \graycell{#6} & \graycell{#7}}
\newcommand{\grayrowB}[6]{& \graycell{#1} & \graycell{#2} & \graycell{#3} & \graycell{#4} & \graycell{#5} & \graycell{#6} \\}

\usepackage{ragged2e}
\begin{document}

\title{
%Physically Plausible Video Generation via Event-Centric CoT with Transition Anchoring and Contrastive Semantics
Physically Plausible Video Generation via Visual-Semantic Chain-of-Events Conditioning
}

\author{Zixuan Wang$^{\dagger}$, Yixin Hu$^{\dagger}$, Wen Li, Feng Chen, Yan Liu, Duo Peng, Yinjie Lei\textsuperscript{\Letter}
        % <-this % stops a space
\thanks{This work was supported by the National Natural Science Foundation of China (No.U23B2013, 62276176). This work was also partly supported by the SICHUAN Provincial Natural Science Foundation (No. 2024NSFJQ0023).}% <-this % stops a space
\thanks{Zixuan Wang, Yixin Hu, and Yinjie Lei are with the College of Electronics and Information Engineering, Sichuan University, Chengdu, 610064, China. E-mail: zixuan980525@gmail.com, yixinhu@stu.scu.edu.cn, yinjie@scu.edu.cn.}
\thanks{Wen Li is with the School of Computer Science and Engineering, University of Electronic Science and Technology of China, Chengdu, 611731, E-mail: liwen@uestc.edu.cn}
\thanks{Feng Chen is with the School of Computer Science, University of Adelaide,
5005, Adelaide, Australia. E-mail: chenfeng1271@gmail.com.}
\thanks{Yan Liu is with the Department of Aeronautical and Aviation Engineering at The Hong Kong Polytechnic University, Hong Kong SAR, China. E-mail: scuiicliuyan@gmail.com}
\thanks{Duo Peng is with the School of Computer Science and Technology, Tongji University, Shanghai, 201804, China, E-mail: duopeng@tongji.edu.cn.}
\thanks{$^{\dagger}$ Equal Contribution.}
\thanks{\textsuperscript{\Letter} Corresponding author: Yinjie Lei 
(e-mail: yinjie@scu.edu.cn)}}

\IEEEtitleabstractindextext{
\begin{abstract}
\justifying
Physically Plausible Video Generation (PPVG) seeks to synthesize videos consistent with physical principles, yet remains challenging due to underspecified natural language conditioning. Advanced chain-of-thought (CoT) frameworks augment prompts with physical knowledge. However, such prompts describe physical phenomena holistically, overlooking intermediate states and transition dynamics. In this paper, we reformulate PPVG as event-centric generation by representing physical evolution as a chain of causally connected and physically constrained events. Our framework comprises three key modules: (1) Physics-driven Event Chain Reasoning. This module decomposes physical phenomena into causally connected events represented by evolving scene graphs. Formula-derived physical quantities are bound to relevant objects and interactions, characterizing the direction and magnitude of each event transition. (2) Transition-aware Routed Keyframe Conditioning. This module routes each event to a specialized keyframe synthesis operator for appearance variation or object transformation. Consecutive keyframes are injected as residual guidance during denoising, enabling smooth visual transitions between event-boundary states. (3) Physics-injected Contrastive Semantic Guidance. This module constructs physics-informed positive and counterfactual negative prompts for classifier-free guidance, steering generation toward plausible dynamics and away from physics-violating counterparts. Experiments on PhyGenBench, VideoPhy, PhyWorldBench, and Physics-IQ demonstrate that our framework generates videos with superior physical plausibility across diverse domains.
\end{abstract}
}

\maketitle
\IEEEdisplaynontitleabstractindextext
\IEEEpeerreviewmaketitle

%\begin{abstract}

%\end{abstract}

%\begin{IEEEkeywords}

%\end{IEEEkeywords}

\begin{figure*}[t]
    \centering
    \includegraphics[width=\textwidth]{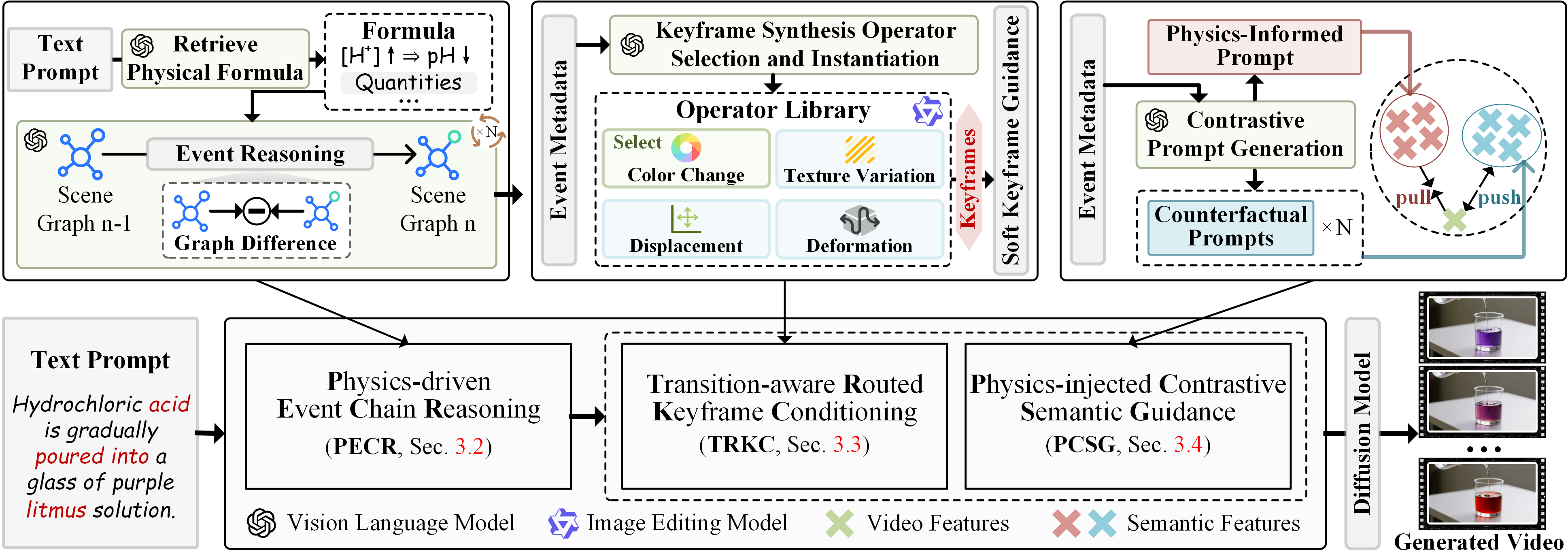}
    \caption{Overview of our physically plausible video generation framework. We begin by inferring a sequence of physically grounded events by reasoning over scene-graph evolution with applicable physical formulas (Section~\ref{sec:3.2}). Then, we route each event to an appropriate editing operator to synthesize a keyframe visually representing the associated physical state transition (Section~\ref{sec:3.3}). In parallel, we derive physics-consistent and counterfactual prompts to provide contrastive semantic guidance (Section~\ref{sec:3.4}). The visual and semantic cues are fed into an off-the-shelf diffusion model to generate videos that faithfully capture the causal progression of physical phenomena.}
    \label{fig:main_figure}
\end{figure*}

\section{Introduction}
\IEEEPARstart{P}{hysically} Plausible Video Generation (PPVG) 
% has opened up 
is valuable for
a wide range of real-world applications, including movie production \cite{zhang2025generative}, autonomous driving simulation \cite{deng2024streetscapes}, and embodied intelligence \cite{agarwal2025cosmos}. Although advanced T2V models \cite{kling2024,liu2024sora} can synthesize photorealistic scenes, % visual realism does not ensure physical plausibility.
they often fail to preserve physical plausibility in the depicted dynamics.
% : a glass may appear realistic yet shatter before impact. 
Thus, current T2V models remain unreliable in reproducing complex physical phenomena. 
  
%Prior
Studies on PPVG have endeavored to improve the physical plausibility of generated videos via graphics engine simulation, feedback-driven optimization, and physics-aware prompt enhancement. Simulation-based approaches \cite{hu2019difftaichi,hu2019taichi,liu2024physgen,hsu2025autovfx,zhang2025physchoreo,tan2026physmotion,foo2026physical} model motion and interactions via graphics engines or differentiable simulators. Feedback-driven approaches \cite{zhang2026physrvg,cai2025phygdpo,yuan2026inference,hassan2026proprio} optimize generation using physics-informs preferences or rewards. Prompt enhancement approaches \cite{yang2025vlipp, zhang2025think,hao2025enhancing,xue2025phyt2v,feng2026newton} employ chain-of-thought (CoT) reasoning to augment user-provided descriptions with physical principles. However, these approaches still struggle to capture the complete causal progression of physical phenomena. Consequently, generated videos remain confined to a single holistic scene snapshot instead of depicting expected evolution.
%, \textit{e.g.}, an ice cube remains solid despite increasing temperature.

We identify three challenges in synthesizing evolving physical phenomena: (1) \textit{Unstructured Causal Modeling.} Physical phenomena unfold as a sequence of causally connected events, each characterized by meaningful physical changes in object configurations instead of a single culminating scene. However, prevailing PPVG approaches \cite{liu2024physgen,zhang2025physchoreo,tan2026physmotion,foo2026physical,zhang2026physrvg,cai2025phygdpo,yuan2026inference,hassan2026proprio,zhang2025think,xue2025phyt2v} embed physical knowledge via video-level supervision or globally augmented descriptions, without clearly specifying how physical states should evolve across events. (2) \textit{Transition-agnostic Visual Anchoring.} Several PPVG approaches provide 
% an opening frame, 
a short video prefix 
% (3$\sim$5 frames) 
\cite{liu2024physgen,zhang2025physchoreo,tan2026physmotion} or a pair of endpoint frames \cite{wan2025wan} as visual evidence. Without dedicated keyframes for successive physical transitions, generative models are compelled to extrapolate unspecified intermediate states, often failing to depict the intervening dynamics. Synthetic optical flow \cite{yang2025vlipp,foo2026physical} provides guidance for object motion, but cannot capture physical phenomena beyond kinematics. (3) \textit{
% Weakly 
Limited Discriminative Physical Semantics.} Physically plausible and implausible dynamics may differ only in subtle yet crucial details, 
% such as motion direction, causal order, or variation magnitude, 
\textit{e.g.}, liquid flows downward versus upward. 
% Therefore, factual semantics provide insufficient cues to distinguish valid dynamics from subtly flawed variants \cite{zhang2025think,xue2025phyt2v}. Likewise, generic negative prompts \cite{pathak2026physvid} merely describe broad undesirable artifacts instead of violations tied to individual events.
However, positive prompts alone provide insufficient cues to distinguish valid dynamics from subtly flawed variants \cite{zhang2025think,xue2025phyt2v}, while generic negative prompts target broad artifacts instead of physical violations associated with individual events \cite{pathak2026physvid}.

To address these challenges, we propose an event-centric CoT framework that couples physics-aware causal reasoning with visual and semantic conditioning, enabling generative models to faithfully depict how physical phenomena unfold over time, as shown in Figure~\ref{fig:main_figure}. Our framework comprises several core modules: (1) We design a \textit{Physics-driven Event Chain Reasoning} (\textit{PECR}) module (Section~\ref{sec:3.2}) to model the causal progression of physical phenomena. 
% This module organizes a complex process as an evolving sequence of physics-aware scene graphs, delineating clear event boundaries (each graph) and causal transitions (graph difference). 
Rather than representing a physical phenomenon with a global description, PECR recursively generates an evolving scene graph chain. Each graph encodes the scene configuration at an event boundary, and differences between consecutive graphs define the corresponding events.
% Also, physical quantities derived analysis of applicable physical formulas are bound to event boundaries, specifying magnitudes and rates of each transition. 
Physical quantities derived from applicable formulas are further  bound to event boundaries, specifying magnitudes and rates of each event transition.
(2) We develop a \textit{Transition-aware Routed Keyframe Conditioning} (\textit{TRKC}) module (Section~\ref{sec:3.3}) to provide dedicated visual anchors for individual intermediate states. To accommodate diverse visual changes induced by events, this module routes each scene transition to an appropriate keyframe editing pathway specialized for appearance changes or object transformations. These keyframes are converted into time-varying soft guidance to bridge the gap between discrete boundary states and continuous video dynamics. (3) We devise a \textit{Physics-njected Contrastive Semantic Guidance} (\textit{PCSG}) module (Section~\ref{sec:3.4}) to enhance the 
%discriminative power 
discriminability
of semantic conditions through 
% contrasting 
contrast between physically plausible dynamics and their implausible counterparts.
% This module condenses event-wise descriptions into a compact positive condition for CFG \cite{ho2022classifier}, preserving salient physical dynamics across events. Besides, it constructs a counterfactual negative condition from perturbations of event-specific physical quantities beyond their physical constraints.
This module condenses event-wise descriptions into a compact positive condition that preserves salient physical dynamics across events. It constructs a counterfactual negative condition from perturbations of event-specific physical quantities beyond their physical constraints. These conditions are used in CFG \cite{ho2022classifier} to provide contrastive semantic guidance for generation.

We evaluate our framework on four benchmarks, including PhyGenBench \cite{meng2024towards}, VideoPhy \cite{bansal2024videophy}, PhyWorldBench \cite{gu2025phyworldbench}, and Physics-IQ \cite{motamed2026generative}. Across diverse video generation backbones, our framework consistently outperforms current PPVG approaches on physics-informed metrics spanning various physical domains and interaction types. Crucially, comprehensive diagnostic analysis demonstrates that videos generated by our framework can preserve both the reasonable chronological order of physical events and the completeness of their intermediate states.

Our contributions are summarized as follows:
\begin{itemize}
\item 
% We propose an event-centric video generation framework that formulates physical phenomena as sequences of causally connected and dynamically evolving events.
We reformulate PPVG as event-centric video generation, representing physical phenomena as chains of causally connected events.
\item 
% To impose structured causal modeling, we construct a scene graph chain with physical quantities to represent event boundaries and transition dynamics.
To impose structured causal modeling, we construct an evolving scene graph chain with formula-derived physical quantities to represent event boundaries and transition dynamics.
\item 
% To visually ground intermediate physical transitions, scene configuration shifts are translated into dedicated visual anchors and propagated as continuous guidance.
To capture intermediate physical transitions, we introduce event-adaptive visual conditioning that bridges discrete event-boundary states and continuous visual evolution.
\item 
% To enhance semantic discriminability, we introduce contrastive physical cues distinguishing admissible event dynamics from physics-violating counterfactual.
To enhance semantic discriminability, we introduce contrastive semantic conditioning that distinguishes plausible event dynamics from corresponding physics-violating counterfactuals.
\end{itemize}

This paper is an extension of our preliminary conference version \cite{wang2026chain} with major improvements: (1) PECR augments scene graph nodes and edges with physical quantities from applicable physical formulas. This provides a shared physical basis for deriving visual and semantic conditions subsequently. (2) For visual conditioning, we redesign the generic keyframe editing pipeline as an adaptively routed architecture comprising local appearance and object transformation. (3) For semantic conditioning, we replace causal narratives with concise descriptions of salient physical changes. Also, we replace manually specified negative prompts with counterfactual ones derived from violations of physical constraints. (4) Beyond PhyGenBench and VideoPhy, we evaluate our framework on PhyWorldBench \cite{gu2025phyworldbench} and Physics-IQ \cite{motamed2026generative} benchmarks, covering broader scenarios. (5) Additional ablation studies are presented.

\section{Related Works}
Prior studies closely associated with our work can be broadly grouped into several categories: physically plausible video generation, CoT in visual generation, and cross-modal conditioning in video generation. These research directions respectively investigate the integration of physical knowledge, the derivation of generation conditions via step-by-step reasoning, and the design of complementary guidance for controllable video synthesis.

\textbf{Physically Plausible Video Generation.} To make videos obey physical laws, physics-aware generation has been increasingly explored. Prior studies generate physically plausible videos \textit{directly from user-provided descriptions} using diffusion models based on 3D U-Net \cite{chen2024videocrafter2,fei2024dysen,bar2024lumiere,wang2025lavie} or Diffusion Transformer (DiT) \cite{liu2024sora,kong2024hunyuanvideo,yang2024cogvideox} architectures. While such studies can render realistic appearance and smooth motion, physical laws remain difficult to model due to the lack of large-scale data with reliable physical annotations. In view of this, several works \cite{hu2019taichi,liu2024physgen,hsu2025autovfx,zhang2025physchoreo,tan2026physmotion,foo2026physical} characterize physical phenomena through simulations based on \textit{graphics engines}, which are further integrated into diffusion sampling to enhance physical realism. Unfortunately, engine parameters require manual specification by users. To handle diverse \textit{open-domain physical phenomena}, VideoREPA \cite{zhang2025videorepa} leverages physical knowledge from foundation models. WISA \cite{wang2025wisa}, PhysHPO \cite{chen2025hierarchical}, and PhysVid \cite{pathak2026physvid} guide diffusion models to learn physical phenomena from decomposed physical principles. MMPhysPipe \cite{lin2026mmphysvideo} and PHANTOM \cite{shen2026phantom} achieve implicit inference of physical properties as a result of jointly learning visual information and physics-aware perceptual cues. 
% To enhance the ability of video generative models to model physical dynamics,
To strengthen the physical modeling capabilities of video generation models,
PhysRVG \cite{zhang2026physrvg} and PhyGDPO \cite{cai2025phygdpo} employ physics-aware reinforcement learning, whereas WMReward \cite{yuan2026inference} and Proprio \cite{hassan2026proprio} apply test-time scaling. To characterize real-world object motion in 3D space, PhysCtrl \cite{wang2026physctrl}, Phys4D \cite{lu2026phys4d}, and OrthoPhys \cite{wang2026physvideo} perform physics-consistent 4D modeling \cite{duan2026liveworld} from video diffusion. VLIPP \cite{yang2025vlipp}, DiffPhy \cite{zhang2025think}, PAG-SAD \cite{hao2025enhancing}, PhyT2V \cite{xue2025phyt2v}, and NEWTON \cite{feng2026newton} leverage CoT reasoning to design physics-aware prompts. However, 
%the above 
these approaches typically incorporate physical knowledge via holistic supervision or globally augmented conditions, without systematically modeling how a physical phenomenon evolves across causally connected events under quantitative physical constraints.   

\textbf{Chain-of-Thought in Visual Generation.} Recent studies have adapted CoT reasoning \cite{wei2022chain} from language understanding to visual generation, achieving fine-grained and interpretable visual synthesis. These studies can be divided into two categories. Some leverage the \textit{reasoning before generation} paradigm to augment conditioning signals. For example, IRG \cite{huang2025interleaving} and Draw-In-Mind \cite{zeng2025draw} refine original descriptions for fine-grained image generation. LayerCraft \cite{zhang2025layercraft} and GoT \cite{fang2025got} enable the generation of multiple objects by reasoning about spatial arrangements. C-Drag \cite{li2025c} infers interactive motion trajectories between objects and their surroundings for drag-driven generation. For image animation, MotiMotion \cite{hsin2026motimotion} hallucinates plausible secondary motions, such as chain reactions resembling domino effects. VChain \cite{huang2025vchain} predicts sparse keyframes to facilitate cross-frame coherent video generation. Others embed step-by-step reasoning into the synthesis process through the \textit{reasoning during generation} paradigm. GVCoT \cite{yin2026generative} generates spatial visual cues within the in-process reasoning steps to localize the region of interest during image editing. At each step of progressive 3D object assembly, SoT \cite{huo2026shape} predicts the structural operation and immediately grounds it by rendering the resulting state. However, visual CoT approaches organize their reasoning process around semantic composition or spatial planning required for the generation. These approaches have only an insufficient ability to perform causal reasoning about physical evolution.

% \textbf{Dual-Prompt in Video Generation.} 
\textbf{Cross-modal Conditioning in Video Generation.} 
While natural language defines scene semantics, it often fails to convey geometry and motion information. In view of this, complementary visual cues are introduced to guide video generation, including reference images, spatial layouts, and motion priors. Some works \cite{girdhar2024factorizing,lin2025stiv,lai2025incorporating,ni2024ti2v,li2023videogen} employ the \textit{reference image} to serve as the appearance prior for generating high-fidelity textures and diverse visual styles. To enhance geometric details, several studies introduce \textit{spatial layouts} during generation. For example, SketchVideo \cite{liu2025sketchvideo} leverages sketches to constrain the contours and structural boundaries of objects. DyST-XL \cite{he2025dyst} and BlobGEN-Vid \cite{feng2025blobgen} specify the locations of multiple objects through bounding boxes and blobs, respectively. CineMaster \cite{wang2025cinemaster} enables the generation of cinematic scenes with realistic parallax by defining camera poses and depth relations in 3D space. Given the dynamic nature of videos, recent studies use \textit{motion priors} to capture sophisticated trajectories. For example, TrackGo \cite{zhou2025trackgo} and Mojito \cite{he2024mojito} regulate the direction and intensity of movement using motion vectors to achieve smooth dynamics. However, 
%many
conditional generation approaches typically specialize in a particular form of visual guidance and therefore have difficulty accommodating heterogeneous changes across various stages of a physical process. 

\section{Methodology}
\subsection{Overall Framework}
Given a user-provided linguistic description $w$ of the physical phenomenon, our goal is to generate the physically plausible video $\mathbf{V}$ that characterizes the underlying progression of the described phenomenon.
%\begin{equation}
%\Gamma: w \rightarrow \mathbf {V}    
%\end{equation}
%where $\Gamma$ denotes our physics-aware video generation framework. 
An overview of our framework is presented in Figure~\ref{fig:main_figure}. Specifically, our framework consists of several synergistic modules. (1) In Section~\ref{sec:3.2}, we design a \textit{Physics-driven Event Chain Reasoning} (\textit{PECR}) module. This module decomposes a complex physical phenomenon into a causally ordered sequence of physical events defined by the evolution of scene graphs. Each event is characterized by both qualitative changes in physical configurations and quantitative estimates of the associated physical quantities. (2) In Section~\ref{sec:3.3}, we develop a \textit{Transition-aware Routed Keyframe Conditioning} (\textit{TRKC}) module. This module selects 
% an appropriate 
a specialized visual synthesis 
% branch
operator
according to the type of visual change induced by each event and progressively generates keyframes aligned with the physical evolution. These keyframes are encoded as smoothly varying soft visual guidance throughout the denoising process. (3) In Section~\ref{sec:3.4}, we devise a \textit{Physics-injected Contrastive Semantic Guidance} (\textit{PCSG}) module. 
% This module condenses an event chain into positive semantic cues of salient physical evolution and constructs unphysical counterfactuals by violating physical constraints. 
This module condenses the event chain into positive semantic cues capturing sailent physical evolution and constructs unphysical counterfactuals from deliberate violations of the underlying physical constraints.
These factual and counterfactual conditions provide contrastive semantic guidance, steering video denoising toward physically plausible progression and away from unphysical dynamics.

\begin{figure*}[t]
    \centering
    \includegraphics[width=\textwidth]{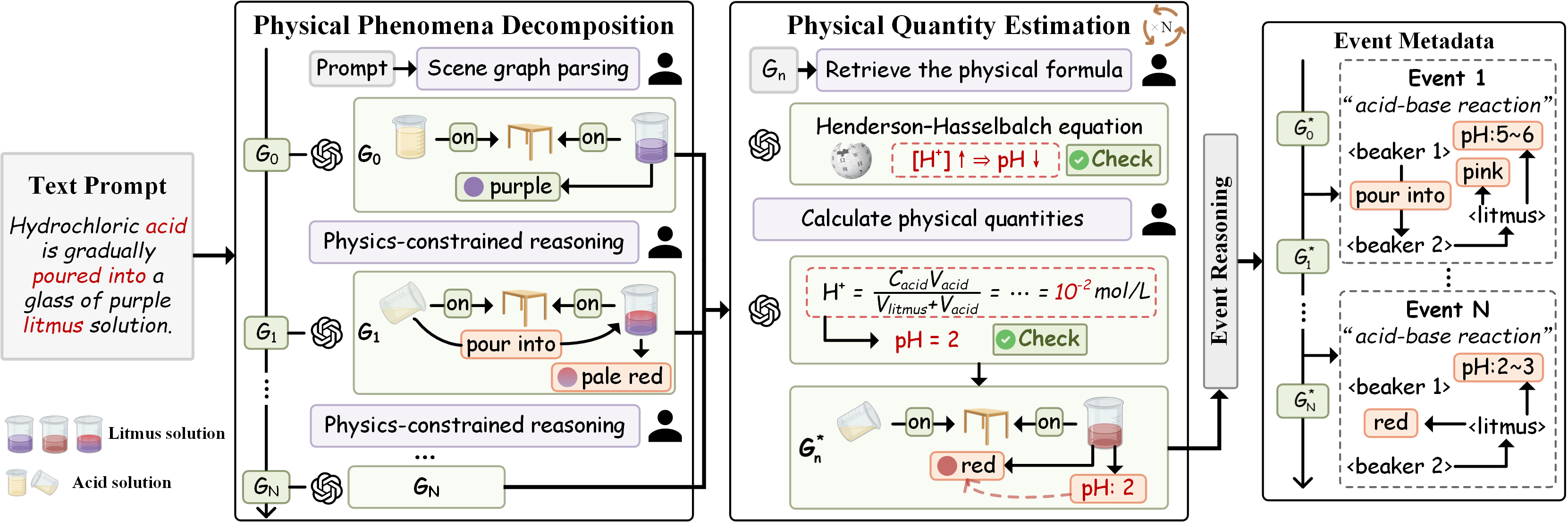}
    \caption{Overview of our proposed PECR module (Section~\ref{sec:3.2}). Given a user-provided prompt, described physical evolution is decomposed into a sequence of scene graphs under inferred physical constraints. Subsequently, applicable physical formulas are searched and applied to calculate physical quantities, augmenting the node and edge properties in scene graphs. Finally, differences between consecutive scene graphs are encoded as metadata of events.}
    \label{fig:module_1}
\end{figure*}

\subsection{Physics-driven Event Chain Reasoning}
\label{sec:3.2}
% A physical phenomenon is better understood as an evolving process involving both qualitative changes in physical configurations and quantitative changes in measurable physical quantities.
In this paper, we characterize physical phenomenon by representing its evolution as a sequence of causally connected events.
Current studies \cite{zhang2025think,xue2025phyt2v} represent a physical phenomenon 
% with a semantic tag centered on the object. 
using an object-centric semantic label.
Such coarse 
% tags
labels
indicate what type of phenomenon occurs (\textit{e.g.}, pouring), but provide limited information about the underlying event structure (\textit{e.g.}, container tilting $\rightarrow$ liquid outflow $\rightarrow$ liquid accumulation). 
% While an LLM can generate a linguistic expansion, this free-form description does not explicitly specify the physical quantities required for quantitative reasoning. 
One straightforward remedy is to prompt an LLM to elaborate coarse semantic labels into detailed descriptions of how the phenomenon unfolds. However, LLMs generally produce free-form descriptions and struggle to determine the values of physical quantities, such as container tilt angles and liquid volumes.
% Therefore, we model physical phenomena by physics-driven event chain reasoning, as shown in Figure~\ref{fig:module_1}. 
Therefore, we characterize the physical phenomena using event chain reasoning constrained by physical formulas, as shown in Figure~\ref{fig:module_1}.
The reasoning pipeline comprises several coupled components. \textit{Physical Phenomena Decomposition} models a physical phenomenon as a causally ordered sequence of scene graphs, defining semantically meaningful boundaries between events. \textit{Physical Quantity Estimation}  grounds individual nodes and edges in the scene graphs to the associated physical quantities and determines their values under formula-based constraints. These augmented scene graphs are finally compared in consecutive pairs to derive descriptors of events.
 
\textbf{Physical Phenomena Decomposition.} We view a complex phenomenon as a sequence of physically meaningful configuration transitions. The transition is defined as an event when it is characterized by observable changes in states of individual objects or relations among multiple objects under physical constraints. 
% Operationally,
In practice,
an LLM is instructed to represent the physical phenomenon as the evolution of scene graphs over time, instead of directly decomposing it into several linguistic event phrases. Each scene graph serves as the scene configuration corresponding to an event boundary. 

Specifically, this module begins by parsing a user-provided linguistic description $w$ into a physics-aware scene graph $\mathcal{G}_{0}=(\mathcal{V}_{0},\mathcal{R}_{0})$. Each node in $\mathcal{V}_{0}$ corresponds to an individual object
% , with its physical states encoded as node attributes. 
together with its observable attributes. 
A relation edge in $\mathcal{R}_{0}$ describes a physical interaction between a pair of objects. Then, this module recursively expands the scene graph sequence via physics-constrained reasoning. Based on $\mathcal{G}_{n-1}$, a collection of physical constraints 
% $\mathcal{P}_{n}$ 
$\mathcal{P}_{n}$ (\textit{e.g.}, ``acid-base reaction”)
relevant to the evolution of the physical phenomenon is inferred. Guided by such constraints, candidate successor scene graphs are generated, % and the most plausible one is selected.
from which $\mathcal{G}_{n}$ is selected as follows:
\begin{equation}
\mathcal{G}_{n}
=
\arg\max_{\hat{\mathcal{G}}_{n}}
\Phi\!\left(\hat{\mathcal{G}}_{n}\mid \mathcal{G}_{n-1},\mathcal{P}_{n}\right),
\end{equation}
where $\hat{\mathcal{G}}_{n}$ denotes a candidate successor scene graph. 
% $\Phi(\cdot)$ denotes a scoring function, ranking candidates under the given physical constraints. 
$\Phi(\cdot)$ denotes a scoring function implemented by an LLM. This function compares $\hat{\mathcal{G}}_{n}$ with $\mathcal{G}_{n-1}$ and assigns a higher score when the resulting changes in the attributes and relations of the object are more consistent with $\mathcal{P}_{n}$.
% This 
The above process yields a causally ordered sequence of scene graphs, 
% $\mathcal{G}_{0} \rightarrow \mathcal{G}_{1} \rightarrow \cdots \rightarrow \mathcal{G}_{N}$.
$\mathcal{G}_{0} \xrightarrow{\mathcal{P}_{1}} \mathcal{G}_{1} \xrightarrow{\mathcal{P}_{2}} \cdots \xrightarrow{\mathcal{P}_{N}} \mathcal{G}_{N}$, with $\mathcal{P}_{n}$ providing the physical basis for the corresponding evolution step.

\textbf{Physical Quantity Estimation.} The above scene graph sequence captures the semantic changes between consecutive boundaries of events, yet the numerical values required to characterize the corresponding dynamics remain unspecified. 
% Therefore, these scene graphs are augmented with a set of physical quantities estimated using physical formulas. 
To provide a quantitative representation, physical formulas are retrieved and used to derive the corresponding physical quantities for each step of evolution. These physical quantities are subsequently assigned to their respective nodes and edges in the scene graphs.

% In particular, an LLM determines formula names $\mathcal{N}_{\mathcal{P}_{n}}$ based on the physical constraint $\mathcal{P}_{n}$, and uses these names to retrieve physical formulas $\mathcal{F}^{*}_{n}$ from knowledge repositories.
Specifically, an LLM generates formula queries $\mathcal{N}_{\mathcal{P}_{n}}$ based on $\mathcal{P}_{n}$, including potential formula names and the physical quantities involved. These queries are used to search online sources for potentially applicable formulas. The obtained formulas are collected into a candidate set $\mathcal{F}_{\mathcal{P}_{n}}$. Afterwards, the single best-matching formula is selected as follows:
%\begin{equation}
% \mathcal{F}^{*}_{n} = \mathrm{TopK}_{f \in \mathcal{F}_{\mathcal{P}_{n}}} \mathrm{sim}(f, [\mathcal{N}_{\mathcal{P}_{n}}, \mathcal{P}_{n}]),
%\end{equation}
\begin{equation}
\mathcal{F}^{*}_{n}
=
\arg\max_{f \in \mathcal{F}_{\mathcal{P}_{n}}}
\mathrm{sim}\!\left(
f,
\mathcal{N}_{\mathcal{P}_{n}}
\right).
\end{equation}
% where $\mathcal{F}_{\mathcal{P}_{n}}$ denotes all formulas associated with physical constraints $\mathcal{P}_{n}$. 
where $\mathrm{sim}(\cdot,\cdot)$ measures the semantic similarity between each candidate physical formula and the corresponding query. If the similarity score of the selected formula $\mathcal{F}_{n}^{*}$ falls below a predefined threshold, the LLM reformulates $\mathcal{N}_{\mathcal{P}_{n}}$ based on the candidate formulas in $\mathcal{F}_{\mathcal{P}_{n}}$, and the selection process is repeated using reformulated query.
% When no direct match of inferred formula names $\mathcal{N}_{\mathcal{P}_{n}}$ is found in $\mathcal{F}_{\mathcal{P}_{n}}$, formula names are regenerated conditioned on $\mathcal{F}_{\mathcal{P}_{n}}$.

\begin{figure*}[t]
    \centering
    \includegraphics[width=\textwidth]{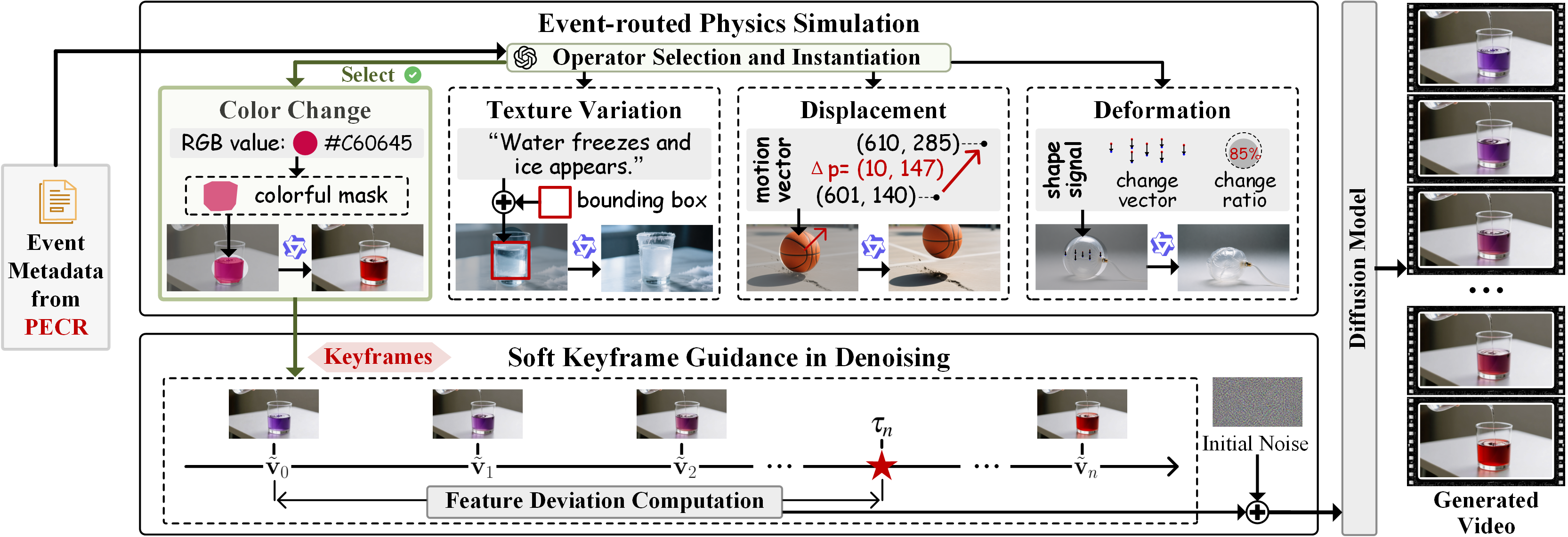}
    \caption{Overview of our proposed TRKC module (Section~\ref{sec:3.3}). For individual events, an appropriate keyframe synthesis operator is selected according to the induced visual change. The selected operator edits the preceding keyframe by modifying object states or scene composition, producing the keyframe after the current event. These keyframes are subsequently used to create smoothly varying visual guidance, which is residually injected during denoising. }
    \label{fig:module_2}
\end{figure*}

Since physical quantities in formulas are typically represented as abstract symbols, they are not directly associated with specific objects encoded in the scene graph. Therefore, after retrieving the formulas $\mathcal{F}^{*}_{n}$, an external symbolic algebra engine \cite{meurer2017sympy} is invoked 
% for deterministic formula parsing.
to parse its variables.
Then, an LLM grounds the parsed symbols to the corresponding nodes and edges in $\mathcal{G}_{n}$. 
% Their numerical values are assigned to these symbols by solving the retrieved formulas or querying online knowledge bases.
Their numerical values are obtained by solving the formula $\mathcal{F}^{*}_{n}$ or querying online knowledge bases.
%\begin{equation}
%\left(\mathcal{V}_n^{*}, \mathcal{R}_n^{*}\right)
%=
%\left(
%\mathcal{V}_n \oplus
%\mathcal{Q}^{V}\!\left(\mathcal{F}_n^{*}\right),
%\mathcal{R}_n \oplus
%\mathcal{Q}^{R}\!\left(\mathcal{F}_n^{*}\right)
%\right),
%\end{equation}
%where $\mathcal{Q}^{\mathcal V}(\mathcal{F}^{*}_{n})$ and $\mathcal{Q}^{\mathcal R}(\mathcal{F}^{*}_{n})$ derive physical quantities from the formulas and compute corresponding numerical values for nodes and edges, respectively. $\oplus$ augments nodes and edges by adding physical quantities, yielding $\mathcal{V}_n^{*}$ and $\mathcal{R}_n^{*}$. 
The scene graph augmented with physical quantities is denoted by $\mathcal{G}_{n}^{*}=(\mathcal{V}_{n}^{*},\mathcal{R}_{n}^{*})$.
% The inferred physical quantities are validated for consistency across consecutive scene graphs. If their changes violate governing physical constraints, the inference is repeated. 
Subsequently, physical quantities are checked across consecutive scene graphs, and recalculated when their changes violate the physical constraints $\mathcal{P}_n$.
%Using the augmented scene graph, each event $e_{n}$ is instantiated as:
Finally, given consecutive augmented scene graphs $\mathcal{G}_{n-1}^{*}$ and $\mathcal{G}_{n}^{*}$, the event $e_n$ is defined as their transition together with the governing physical constraints $\mathcal{P}_n$:
%\begin{equation} 
%e_n = \Delta\!\left( \mathcal{G}_{n-1}^{*}, \mathcal{G}_n^{*} \right), 
%\end{equation} 
\begin{equation}
e_n =
\left\{
\mathcal{G}_{n-1}^{*} \rightarrow \mathcal{G}_{n}^{*},
\mathcal{P}_{n}
\right\},
\end{equation}
% where $\Delta(\cdot, \cdot)$ identifies changes in nodes or edges. 
where $\mathcal{G}_{n-1}^{*} \rightarrow \mathcal{G}_{n}^{*}$ represents qualitative and quantitative changes in objects and their relations.
% Finally, a physics-aware scene graph sequence $\mathcal{G}_0^{*} \xrightarrow{e_1} \mathcal{G}_1^{*} \xrightarrow{e_2} \cdots \xrightarrow{e_N} \mathcal{G}_N^{*}$ is produced.
The augmented scene graphs and the events defined between consecutive pairs constitute the event chain, $\mathcal{G}_0^{*} \xrightarrow{e_1} \mathcal{G}_1^{*} \xrightarrow{e_2} \cdots \xrightarrow{e_N} \mathcal{G}_N^{*}$.

\subsection{Transition-aware Routed Keyframe Conditioning}
\label{sec:3.3}
This module maps the physical evolution represented by the event chain into visual guidance for video generation, as shown in Figure~\ref{fig:module_2}. Formally, inferred event chain is visually grounded into a keyframe sequence $\mathbf{v}_0 \xrightarrow{e_1} \mathbf{v}_1 \xrightarrow{e_2} \mathbf{v}_2 \cdots \xrightarrow{e_N} \mathbf{v}_N$. Conditioned on $\mathcal{G}^{*}_{0}$, $\mathbf{v}_0$ is generated using a T2I model. Each following keyframe $\mathbf{v}_n$ is progressively edited from $\mathbf{v}_{n-1}$ according to $e_{n}$. These keyframes are subsequently converted into smoothly varying visual guidance for video generation. The pipeline above comprises two sequential components responsible for keyframe synthesis and visual guidance, respectively. For keyframe synthesis, \textit{Event-Routed Physics Simulation} routes each event to a specialized simulation operator according to its induced physical change, producing event-boundary keyframes that faithfully capture changes in object appearance, spatial configuration, or geometry. For visual guidance, \textit{Soft Keyframe Guidance in Denoising} derives smoothly varying features from synthesized keyframes and injects them residually during denoising process.

% \textbf{Event-Routed Physics Simulation.} Events manifest as heterogenous changes in object appearance, spatial configuration, and geometry, making a uniform editing scheme insufficient. To accommodate such diversity, each event $e_n$ is routed to the corresponding branch in a bank of specialized physics simulation operators. This maps the specified physical change into spatially grounded conditions for transforming keyframe $\mathbf{v}_{n-1}$ to $\mathbf{v}_{n}$. Before applying the selected operator, SAM \cite{kirillov2023segment} segments the region potentially affected by $e_n$ in $\mathbf{v}_{n-1}$, providing the spatial constraint for the editing. The specialized operators and default operation are described below.
\textbf{Event-routed Physics Simulation.} Physical events manifest as heterogeneous changes in object appearance, spatial configuration, and geometry, making a uniform editing scheme insufficient. To accommodate such diversity, each event $e_n$ is routed to a specialized physics simulation operator depending on its induced visual effect. Given the scene transition $\mathcal{G}^{*}_{n-1}\rightarrow\mathcal{G}^{*}_n$ contained in $e_n$ and the preceding keyframe $\mathbf{v}_{n-1}$, the VLM infers the spatial cues (\textit{e.g.}, color values, deformation parameters) and the semantic editing conditions required by the selected operator. 
\begin{equation}
(\hat{\mathcal{O}}_{{\rm img},n}, \mathcal{O}_{{\rm txt},n}) = \mathrm{VLM}\big(\mathcal{G}^{*}_{n-1} \rightarrow  \mathcal{G}^{*}_n; \mathbf{v}_{n-1}\big),
\end{equation}
where $\hat{\mathcal{O}}_{{\rm img},n}$ and $\mathcal{O}_{{\rm txt},n}$ denote the spatial cue and the semantic editing condition, respectively. Then, SAM segments the affected region in $\mathbf{v}_{n-1}$, yielding a mask that spatially constrains the edit. The spatial cue is subsequently rendered onto such a masked region to produce a conditioning image $\mathcal{O}_{{\rm img},n}$. Finally, Qwen-Image-Edit \cite{wu2025qwen} is employed to generate keyframe $\mathbf{v}_{n}$.
\begin{equation}
\mathbf{v}_n = \mathrm{Edit}(\mathbf{v}_{n-1}; \mathcal{O}_{{\rm img},n}, \mathcal{O}_{{\rm txt},n}).
\end{equation} 
While sharing a unified formulation, physics simulation operators differ in how the condition pair $(\mathcal{O}_{{\rm img},n}, \mathcal{O}_{{\rm txt},n})$ is configured for diverse visual changes. 

Accordingly, the operator set comprises four specialized operators for \textit{color change}, \textit{texture variation}, \textit{displacement}, and \textit{deformation}, together with a default operator for the remaining cases. (1) \textit{Color Change}. VLM is used to infer a target RGB value from $\mathcal{\hat{V}}_{n-1:n}^{*}$ and $\mathbf{v}_{n-1}$. The inferred RGB value is assigned to pixels within the segmented region to yield a localized color prior. This guides the editor to update the segmented region toward the inferred color, while preserving its geometry. (2) \textit{Texture Variation}. This variation often emerges progressively with varying spatial coverage and intensity, \textit{e.g.}, frost may initially emerge locally before expanding over the surface. Thus, VLM predicts both a bounding box indicating the spatial extent of the variation and a brief description of the desired surface appearance. The intersection of the bounding box and segmented region defines the editing area. Within this area, the inferred description guides the localized texture generation. (3) \textit{Displacement}. The segmented object is isolated from $\mathbf{v}_{n-1}$ as a movable foreground layer. Based on $\mathcal{\hat{R}}_{n-1:n}^{*}$ and $\mathbf{v}_{n-1}$, VLM estimates the translation and rotation of the foreground object in the image plane. The object is then moved to the estimated position and orientation. Editor completes the vacated region and blends the moved object into the surrounding scene. (4) \textit{Deformation}. The VLM parameterizes geometric changes using quantities such as scale ratio and contour offset, while also describing the deformation type and tendency. These geometric parameters are used to redraw a deformed object mask. And, the editor reconstructs the object within this mask according to the deformation semantics. (5) \textit{Default}. Events beyond the scope of the above specialized simulation operators are handled by inserting or removing the affected scene elements.

\begin{figure*}[t]
    \centering
    \includegraphics[width=\textwidth]{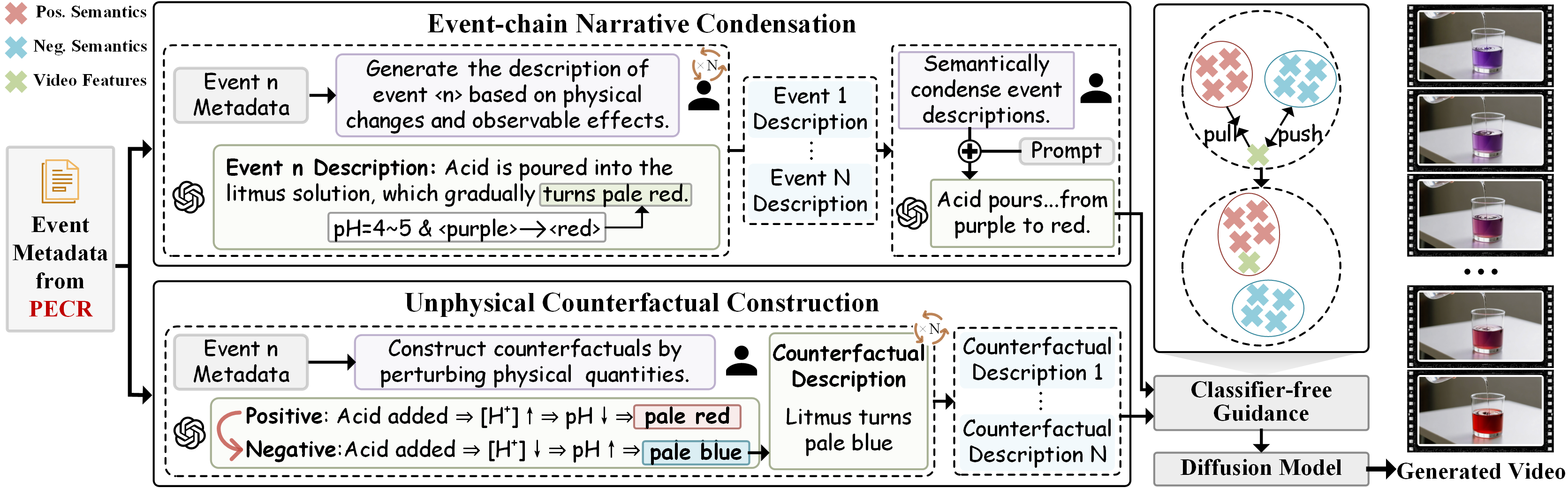}
    \caption{Overview of our proposed PCSG module (Section~\ref{sec:3.4}). The descriptions of individual events are condensed into a compact positive semantic condition by preserving physical changes and observable effects. In parallel, counterfactual negative semantic conditions are derived by perturbing physical quantities of individual events. Both conditions are simultaneously fed into classifier-free guidance for physically plausible video generation.}
    \label{fig:module_3}
\end{figure*}

\textbf{Soft Keyframe Guidance in Denoising.} Once the ordered keyframe sequence is produced, each keyframe is encoded into the compressed feature space as $\widetilde{\mathbf{v}}_{n}=\psi_{\text{img}}(\mathbf{v}_{n})$, where $\psi_{\text{img}}(\cdot)$ denotes the VAE encoder. To provide a smooth progression, linear interpolation is applied between adjacent keyframes along the temporal dimension.. 
\begin{equation}
\widetilde{\mathbf{v}}_{\tau_n} = (1-\lambda_{\tau_n})\widetilde{\mathbf{v}}_{n-1} + \lambda_{\tau_n}\widetilde{\mathbf{v}}_n, \qquad \lambda_{\tau_n}\in[0,1],
\end{equation}
where $\tau_n$ denotes a position between successive keyframes $\widetilde{\mathbf{v}}_{n-1}$ and $\widetilde{\mathbf{v}}_n$, and $\lambda_{\tau_n}$ is the corresponding interpolation ratio. Instead of directly adopting the interpolated feature $\mathbf{z}_{\tau_n}$ as soft guidance, this component employs its deviation from initial keyframe representation $\widetilde{\mathbf{v}}_{0}$.
\begin{equation}
\begin{aligned}
\mathbf{d}_{\tau_n}
&=
\widetilde{\mathbf{v}}_{\tau_n}
-
\widetilde{\mathbf{v}}_0 \\
&=
\sum_{m=0}^{n-2}
\left(
\widetilde{\mathbf{v}}_{m+1}
-
\widetilde{\mathbf{v}}_m
\right)
+
\lambda_{\tau_n}
\left(
\widetilde{\mathbf{v}}_n
-
\widetilde{\mathbf{v}}_{n-1}
\right),
\end{aligned}
\end{equation}
where $\mathbf{d}_{\tau_n}$ denotes the feature deviation at position $\tau_n$. This formulation accumulates the completed variations across preceding keyframes, together with the ongoing change between $\widetilde{\mathbf{v}}_{n-1}$ and $\widetilde{\mathbf{v}}_n$. During denoising, $\mathbf{d}_{\tau_n}$ is residually injected into the corresponding latent.
\begin{equation}
\widetilde{\mathbf{z}}_{\tau_n}
=
\mathbf{z}_{\tau_n}
+
\beta\mathbf{d}_{\tau_n},
\end{equation}
where $\mathbf{z}_{\tau_n}$ and $\widetilde{\mathbf{z}}_{\tau_n}$ denote the noisy features before and after residual guidance, respectively. Coefficient $\beta$ scales the guidance strength and gradually increases as denoising proceeds. Finally, a series of guided features $\{\widetilde{\mathbf{z}}_{1},\widetilde{\mathbf{z}}_{2},\ldots, \widetilde{\mathbf{z}}_{N}\}$ is concatenated along the sequence dimension to form the updated noisy video feature $\widetilde{\mathbf{Z}}$.

\subsection{Physics-injected Contrastive Semantic Guidance}
\label{sec:3.4}
Simply verbalizing $\mathcal{G}^{*}_{0} \xrightarrow{e_{1}} \mathcal{G}^{*}_{1} \xrightarrow{e_{2}} \cdots \xrightarrow{e_{N}} \mathcal{G}^{*}_{N}$ may produce duplicative information and dilute the salient dynamic semantics. In addition, a single positive description specifies what should happen but not which physically implausible dynamics should be avoided. Given this, as shown in Figure~\ref{fig:module_3}, \textit{Event-Chain Narrative Condensation} is proposed to consolidate a series of event-wise descriptions into a compact sentence $\mathcal{W}^{*}_{+}$, capturing critical physical evolutions for generation. \textit{Unphysical Counterfactual Construction} deliberately violates the physical laws governing each event to derive counterfactual cues $\mathcal{W}^{*}_{-}$, guiding video generation away from physically implausible content. 
% During generation, $\mathcal{W}^{*}_{+}$ and $\mathcal{W}^{*}_{-}$ are embedded separately. They are utilized as positive and negative semantic conditions, respectively, via the classifier-free guidance \cite{ho2022classifier} paradigm.
During video generation, $\mathcal{W}^{*}_{+}$ and $\mathcal{W}^{*}_{-}$ are encoded separately and used as  positive and negative conditions for classifier-free guidance \cite{ho2022classifier}.
\begin{equation}
\mathbf{W}_{+}=\psi_{\mathrm{txt}}(\mathcal{W}^{*}_{+}),
\qquad
\mathbf{W}_{-}=\psi_{\mathrm{txt}}(\mathcal{W}^{*}_{-}),
\end{equation}
\begin{equation}
\begin{aligned}
\hat{\boldsymbol{\epsilon}}_{\theta}
={}&
\boldsymbol{\epsilon}_{\theta}
\left(
\mathbf{\widetilde{Z}}_{\tau_z}, \tau_z, \mathbf{W}_{-}
\right)
\\
&+
\gamma
\Bigl[
\boldsymbol{\epsilon}_{\theta}
\left(
\mathbf{\widetilde{Z}}_{\tau_z}, \tau_z, \mathbf{W}_{+}
\right)
-
\boldsymbol{\epsilon}_{\theta}
\left(
\mathbf{\widetilde{Z}}_{\tau_z}, \tau_z, \mathbf{W}_{-}
\right)
\Bigr],
\end{aligned}
\end{equation}
where $\psi_{\text{txt}}(\cdot)$ denotes the text encoder. $\mathbf{W}_{+}$ and $\mathbf{W}_{-}$ denote positive and negative text embeddings, respectively. $\widetilde{\mathbf{Z}}_{\tau_z}$ denotes the noisy video latent representation at diffusion timestep $\tau_z$. $\boldsymbol{\epsilon}_{\theta} (\mathbf{\widetilde {Z}}_{\tau_z},\tau_z,\mathbf{W})$ denotes the conditional noise prediction produced by the denoising network parameterized by $\theta$. $\hat{\boldsymbol{\epsilon}}_{\theta}$ denotes guided noise prediction, and $\gamma$ denotes classifier-free guidance scale.

\textbf{Event-chain Narrative Condensation.} To enhance a user-provided description $w$ with physics-aware dynamics, the inferred scene graph sequence is converted into compact semantic cues describing salient physical evolutions. For each pair of scene graphs, observable effects of objects are verbalized via comparing semantic differences in nodes and edges. Instead of directly inserting numerical values, physical quantities are converted into coarse degree modifier, \textit{e.g.}, ``pale”, defining the corresponding physical changes. Foreground objects and background information remaining unchanged are excluded to avoid redundant descriptions. Each $e_{n}$ is described independently, without using ordering expressions 
% (\textit{e.g.}, ``beforehand”, ``afterward”) 
or adding information absent from scene graphs. 
%\begin{equation}
%w^{*}_{+,n}
%=
%\mathrm{Verb}
%\left(
%\left(
%\mathcal{V}_{n-1}^{*},
%\mathcal{R}_{n-1}^{*}
%\right)
%\rightarrow
%\left(
%\mathcal{V}_{n}^{*},
%\mathcal{R}_{n}^{*}
%\right)
%\right),
%\end{equation}
%where $w_n$ denotes the verbalization of $e_n$. $\mathrm{Verb}(\cdot)$ denotes the verbalization operator.
Video diffusion models typically condition on a single sentence. However, simply assembling several descriptions may produce semantic interference.  Given this, $w^{*}_{+,1}, w^{*}_{+,2}, \ldots, w^{*}_{+,N}$ are fused into a concise global description via semantic condensation of duplicate object references and overlapping physical changes. Finally, this condensed description is combined with a user-provided one to form the physics-enhanced semantic cues $\mathcal{W}^{*}_{+}$ for video generation. 

\textbf{Unphysical Counterfactual Construction.} Alongside semantic cues describing the evolution of a physical phenomenon, corresponding counterfactual descriptions violating physical plausibility of individual events are produced.  For each $e_{n}$, an LLM determines a minimal set of physically significant changes in the nodes or edges between $\mathcal{G}^{*}_{n-1}$ and $\mathcal{G}^{*}_{n}$ according to physical constraints $\mathcal{P}_{n}$. Afterwards, a counterfactual sentence $ w^{*}_{-,n}$ is obtained by inferring how the physical evolution would manifest under a violation of $\mathcal{P}_{n}$, while keeping the involved objects and scene background unchanged. Such violations are instantiated via intervening on physical quantities, because quantities provide an actionable interface between abstract physical laws and observable event dynamics. For example, ``liquid flows downward from tilted container” is recast as the counterfactual `` liquid flows upward along tilted container” by reversing the velocity direction.
%\begin{equation}
%w^{*}_{-,n}
%=
%\operatorname{Verb}\!\left(
%\left(
%\mathcal{V}_{n-1}^{*},
%\mathcal{R}_{n-1}^{*}
%\right)
%\rightarrow
%\left(
%\mathcal{V}_{n}^{*},
%\mathcal{R}_{n}^{*}
%\right)
%\;\middle|\;
%\neg \mathcal{P}_{n}
%\right),
%\end{equation}
%where $\neg \mathcal{P}_{n}$ denotes the counterfactual violation of the physical constraints $\mathcal{P}_{n}$. 
Each $w^{*}_{-,n}$ is generated independently for a single event, without modifying the original scene graph sequence or propagating the induced violation to neighboring events. These sentences collectively form a counterfactual set $\mathcal{W}^{*}_{-} =\left\{w^{*}_{-,1}, w^{*}_{-,2}, \ldots, w^{*}_{-,N} \right\}$, providing negative semantic guidance for video generation.

\section{Experiments}
\subsection{Experimental Setups}
This section sequentially describes the datasets and evaluation metrics used in our experiments, as well as implementation details of our approaches.

\indent %\textbf{Datasets.} 
\textbf{Benchmarks.} The datasets used in our study are: (1) PhyGenBench \cite{meng2024towards}. This encompasses 160 linguistic descriptions spanning 27 physical laws across four fundamental domains, namely “mechanics”, “optics”, “thermal”, and “material”. (2) VideoPhy \cite{bansal2024videophy}. This provides 688 human-verified linguistic prompts, describing various physical interactions between objects, comprising “solid-solid”, “solid-fluid”, and “fluid-fluid”. (3) PhyWorldBench \cite{gu2025phyworldbench}. We adopt the fundamental physical categories of PhyWorldBench, including “object motion and kinematics”, “interaction dynamics”, “deformations and elasticity”, “energy conservation”, “fluid and particle dynamics”, and “lighting and shadow”. Additionally, we use an anti-physics category, where prompts deliberately violate real-world physics. 
% Each category is organized into 5 subcategories. 7 scenarios are defined for each subcategory. For each scenario, 3 prompts are provided, offering diverse levels of informativeness. 
(4) Physics-IQ \cite{motamed2026generative}. 
% This comprises 396 videos (8s), covering 66 distinct physical scenarios. 
This benchmark comprises 396 8s videos covering 66 distinct physical scenarios. 
% The video is split into a 3s conditioning segment and a 5s GT continuation. 
Each video consists of a 3s conditioning segment followed by a 5s GT continuation.
% V2V models are conditioned on the full 3s segment, whereas I2V models are conditioned only on its final frame. 
For I2V evaluation, only the final frame of the 3s conditioning segment is used as input.
% Each scenario is designed around a specific physical law to evaluate the understanding of a video generative model of physical phenomena.
The 66 scenarios are designed around specific physical laws to evaluate a video generative model’s understanding of physical phenomena.

\textbf{Evaluation Metrics.} The evaluation metrics used in our study are as follows. (1) Following PhyGenBench \cite{meng2024towards}, we use Physical Commonsense Alignment (PCA) as our metric, which indicates video quality by considering key phenomena detection, physics order verification, and overall naturalness evaluation. (2) For VideoPhy \cite{bansal2024videophy}, we use its VideoCon-Physics evaluator to assess Semantic Adherence (SA) and Physical Commonsense (PC). SA evaluates whether a linguistic description is semantically grounded in generated video frames. PC examines whether the depicted actions and object properties conform to real-world physics laws. (3) In PhyWorldBench \cite{gu2025phyworldbench}, generated videos are evaluated against predefined Basic Standards and Key Standards.
%a Yes/No evaluation metric is used to assess whether a model can generate accurate videos. Two evaluation standards are defined: Basic Standards and Key Standards. 
%Basic Standards specify the essential objects to be included in the video and require the main actions to be depicted. Key Standards describe the key physical phenomena expected to occur. Following VideoPhy \cite{bansal2024videophy}, SA and PC are used to evaluate video quality. SA checks whether objects align with the video. PC assesses whether each Key Standard is satisfied. 
Basic Standards specify the essential objects and main actions to be depicted, whereas Key Standards describe the expected physical phenomena. SA and PC measure compliance with the Basic Standards and Key Standards, respectively.
(4) According to Physics-IQ \cite{motamed2026generative}, we use Spatial IoU to evaluate where the action happens, Spatiotemporal IoU to measure where and when the action happens, Weighted Spatial IoU to verify where and to what extent the action happens, Mean Squared Error (MSE) to examine assess how the action unfolds. These four metrics are combined into a single Physics-IQ score by summing their individual values, applying a negative sign to the MSE term since lower values indicate better performance.

\begin{table}[htbp]
\centering
\caption{Performance comparison on PhyGenBench \cite{meng2024towards} across four physical domains. The best and second-best results are \textbf{highlighted} and \underline{underlined}, respectively. 
% $^{\dagger}$ indicates cross-benchmark (VideoPhy-2 \cite{bansal2025videophy} $\rightarrow$ PhyGenBench \cite{meng2024towards}) generalization.
For consistency, all PCA scores are reported as percentages with one decimal place.
}
\small
\resizebox{\columnwidth}{!}{%
\begin{tabular}{lccccc}
\toprule
\multirow{2}{*}{Methods} & \multicolumn{4}{c}{Physical domains (\%)} & \multirow{2}{*}{Avg.} (\%) \\
\cmidrule{2-5}
 & Mechanics & Optics & Thermal & Material &  \\
\midrule
\midrule
\multicolumn{6}{l}{\textit{Video Foundation Model}} \\
\textcolor{black}{Lavie \cite{wang2025lavie}} & \textcolor{black}{30.0} & \textcolor{black}{44.0} & \textcolor{black}{38.0}
& \textcolor{black}{32.0} & \textcolor{black}{36.0} \\

\textcolor{black}{VideoCrafter v2.0 \cite{chen2024videocrafter2}} & - & - & - & - & \textcolor{black}{48.0} \\

\textcolor{black}{Open-Sora v1.2 \cite{zheng2024open}} & \textcolor{black}{43.0} & \textcolor{black}{50.0} & \textcolor{black}{34.0}
& \textcolor{black}{37.0} & \textcolor{black}{44.0} \\

\textcolor{black}{Vchitect v2.0 \cite{fan2025vchitect}} & \textcolor{black}{41.0} & \textcolor{black}{56.0} & \textcolor{black}{44.0}
& \textcolor{black}{37.0} & \textcolor{black}{45.0} \\

\textcolor{black}{Kling \cite{kling2024}} & \textcolor{black}{45.0} & \textcolor{black}{58.0} & \textcolor{black}{50.0}
& \textcolor{black}{40.0} & \textcolor{black}{49.0} \\

\textcolor{black}{Pika \cite{Pika2024}} & \textcolor{black}{35.0} & \textcolor{black}{56.0} & \textcolor{black}{43.0}
& \textcolor{black}{39.0} & \textcolor{black}{44.0} \\

\textcolor{black}{Gen-3 \cite{runway2024gen3alpha}} & \textcolor{black}{45.0} & \textcolor{black}{57.0} & \textcolor{black}{49.0}
& \textcolor{black}{51.0} & \textcolor{black}{51.0} \\

\midrule
\midrule
\multicolumn{6}{l}{\textit{Physics-aware Video Generation Model}} \\
WISA \cite{wang2025wisa}                         & -    & -    & -    & -    & 43.0 \\
DiffPhy \cite{zhang2025think}                    & 53.0 & 59.0 & 58.0 & 46.0 & 54.0 \\ 
VideoDPO \cite{liu2025videodpo}                  & 48.0 & 60.0 & 47.0 & 58.0 & 54.0 \\
%LTX-Video-2B$^{\dagger}$ \cite{hacohen2024ltx}               & 0.51 & 0.58 & 0.48 & 0.45 & 0.51 \\
%\, + NEWTON$^{\dagger}$ \cite{feng2026newton}                & 0.50 & 0.65 & 0.62 & 0.54 & 0.56 \\
CogVideoX-5B \cite{yang2024cogvideox}            & 39.0 & 55.0 & 40.0 & 42.0 & 45.0 \\
\, + PhyT2V \cite{xue2025phyt2v}                 & 45.0 & 55.0 & 43.0 & 53.0 & 50.0 \\
\, + SGD \cite{hao2025enhancing}                 & 49.0 & 58.0 & 42.0 & 48.0 & 49.0 \\
%\, + Vanilla DPO \cite{Wallace2024DiffusionDPO}  & 0.48 & 0.60 & 0.47 & 0.58 & 0.54 \\
\, + PhysHPO \cite{chen2025hierarchical}         & 55.0 & 68.0 & 50.0 & \underline{65.0} & 61.0 \\
\rowcolor{DarkGray}
\, + Ours          & \underline{70.0} & \textbf{78.7} & \textbf{76.7} & 64.2 & \textbf{72.5} \\ 
\midrule
\textcolor{black}{Wan2.1-1.3B-VACE \cite{wan2025wan}} & 50.0 & 63.0 & 51.0 & 46.0 & 53.0 \\
\rowcolor{DarkGray}
\, + Ours & 60.8 & 72.7 & 61.1 & 55.8 & 63.3 \\
\midrule
\textcolor{black}{Wan2.1-14B \cite{wan2025wan}} & \textcolor{black}{36.0} & \textcolor{black}{53.0} & \textcolor{black}{36.0}
& \textcolor{black}{33.0} & \textcolor{black}{40.0} \\
\, + PhyGDPO \cite{cai2025phygdpo} & 55.0 & 60.0 & 58.0 & 47.0 & 55.0 \\
\rowcolor{DarkGray}
\, + Ours   & \textbf{70.8} & \underline{78.0} & \underline{71.1} & 61.7 & 70.8 \\
\midrule
\textcolor{black}{Wan2.2-14B \cite{wan2025wan}}  & 53.0 & 61.0 & 58.0 & 43.0 & 54.0 \\
\rowcolor{DarkGray}
\, + Ours  & \underline{70.0} & 76.0 & \textbf{76.7} & \textbf{65.8} & \underline{72.1} \\

\bottomrule
\end{tabular}%
}
\label{tab:phygenbench}
\end{table}

\begin{table*}[htbp]
\centering
\caption{Performance comparisons on VideoPhy \cite{bansal2024videophy} across various physical interactions between objects. The best and second-best results are \textbf{highlighted} and \underline{underlined}, respectively.}
\small
\begin{tabularx}{\textwidth}{l*{12}{Y}}
\toprule
\multirow{2}{*}{Methods} & \multicolumn{3}{c}{Overall (\%)} & \multicolumn{3}{c}{Solid-Solid (\%)} & \multicolumn{3}{c}{Solid-Fluid (\%)} & \multicolumn{3}{c}{Fluid-Fluid (\%)} \\
\cmidrule(lr){2-4}\cmidrule(lr){5-7}\cmidrule(lr){8-10}\cmidrule(lr){11-13}
& \SAPC & SA & PC & \SAPC & SA & PC & \SAPC & SA & PC & \SAPC & SA & PC \\
\midrule
\midrule
\multicolumn{13}{l}{\textit{Video Foundation Model}} \\
\grayrowA{VideoCrafter2 \cite{chen2024videocrafter2}}{19.0}{48.5}{34.6}{4.9}{31.5}{23.8}
\grayrowB{27.4}{57.5}{41.8}{32.7}{69.1}{43.6}

\grayrowA{LaVIE \cite{wang2025lavie}}{15.7}{48.7}{28.0}{8.5}{37.3}{19.0}
\grayrowB{15.8}{52.1}{30.8}{34.5}{69.1}{43.6}

\grayrowA{SVD-T2I2V \cite{blattmann2023stable}}{11.9}{42.4}{30.8}{4.2}{25.9}{27.3}
\grayrowB{17.1}{52.7}{32.9}{18.2}{58.2}{34.5}

\grayrowA{ZeroScope \cite{cerspenseZeroscope576w2023}}{11.9}{30.2}{32.6}{6.3}{17.5}{22.4}
\grayrowB{14.4}{40.4}{37.0}{20.0}{36.4}{47.3}

\grayrowA{OpenSora \cite{zheng2024open}}{4.9}{18.0}{23.5}{1.4}{7.7}{23.8}
\grayrowB{7.5}{30.1}{21.9}{7.3}{12.7}{27.3}

OmniVDiff \cite{xi2026omnivdiff} & - & 45.1 & 11.0 & - & 31.5 & 4.9 & - & 49.3 & 12.3 & - & 69.1 & 23.6 \\

CogVideoX-2B \cite{yang2024cogvideox}            & 18.6 & 47.2 & 34.1 & 12.7 & 42.9 & 28.1 & 21.9 & 56.1 & 34.9 & 25.4 & 34.5 & 47.2 \\

\grayrowA{Pika \cite{Pika2024}}{19.7}{41.1}{36.5}{13.6}{24.8}{36.8}
\grayrowB{16.3}{46.5}{27.9}{44.0}{68.0}{58.0}

\grayrowA{Dream Machine \cite{LumaDreamMachine2024}}{13.6}{61.9}{21.8}{12.6}{50.0}{24.3}
\grayrowB{16.6}{68.1}{23.6}{9.0}{76.3}{11.0}

\grayrowA{Lumiere \cite{bar2024lumiere}}{9.0}{38.4}{27.9}{8.4}{26.6}{27.3}
\grayrowB{9.6}{47.3}{26.0}{9.1}{45.5}{34.5}

\grayrowA{Gen-2 \cite{esser2023structure}}{7.6}{26.6}{27.2}{4.0}{8.9}{37.1}
\grayrowB{8.1}{38.5}{18.5}{15.1}{37.7}{26.4}

\midrule
\midrule
\multicolumn{13}{l}{\textit{Physics-aware Video Generation Model}} \\
CogVideoX-5B \cite{yang2024cogvideox}            
& 30.9 & 63.3 & 39.9 & 18.8 & 50.3 & 33.5 & 41.3 & 76.5 & \textbf{44.1} & 34.5 & 61.8 & 45.4 \\
\, + PhyT2V \cite{xue2025phyt2v}                 & 40.1 & - & - & 25.4 & - & - & 48.6 & - & - & 55.4 & - & - \\
\, + VideoREPA \cite{zhang2025videorepa}         & -    & 49.5  & 29.9 & 22.3 & - & - & 32.2 & - & - & 22.3 & - & - \\
% \, + MMPhysVideo \cite{lin2026mmphysvideo} 
\, + PhysVideo \cite{lin2026mmphysvideo} 
& - & 72.7 & 41.0 & -    & 60.8 & 28.7 & - & 81.5 & 41.8 & - & 80.0 & 41.0  \\
\, + Vanilla DPO \cite{Wallace2024DiffusionDPO}  & 41.3 & - & - & 28.2 & - & - & 50.0 & - & - & 51.8 & - & - \\
\, + PhysHPO \cite{chen2025hierarchical}      & 45.9 & - & - & 32.4 & - & - & 54.1 & - & - & 58.9 & - & - \\
\rowcolor{DarkGray}
\, + Ours         & \textbf{58.5} & \textbf{74.9} & \textbf{42.0} & \textbf{50.0} & \textbf{61.9} & \textbf{38.1} & \textbf{63.0} & \textbf{82.6} & \underline{43.2} & \textbf{63.8} & \textbf{81.1} & \underline{46.5} \\
\midrule
Wan2.1-1.3B-VACE \cite{wan2025wan} & - & 68.0 & 24.4 & - & 60.1 & 16.8 & - & 72.6 & 25.3 & - & 76.4 & 41.8 \\
% \, + MMPhysVideo \cite{lin2026mmphysvideo} 
\, + PhysVideo \cite{lin2026mmphysvideo} 
& - & 69.5 & 27.3 & - & 60.1 & 17.5 & - & 76.7 & 29.5 & - & 74.5 & 47.3 \\
\rowcolor{DarkGray}
\, + Ours  &51.3 &69.8 &32.9 &42.5 &60.2 &24.8 &55.9 &76.9 &34.9 &61.0 &74.6 & \textbf{47.4} \\
\midrule
Wan2.2-14B \cite{wan2025wan} & 39.2 & 48.5 & 30.0 & 33.7 & 40.2 & 27.1 & 45.4 & 60.1 & 30.7 & 37.4 & 39.3 & 35.5 \\
\rowcolor{DarkGray}
\, + Ours  &\underline{57.3} &\underline{73.5} & \underline{41.1} &\underline{49.4} &\underline{61.1} & \underline{37.8} &\underline{61.8} &\underline{81.7} &42.0 &\underline{62.8} &\underline{80.3} &45.3 \\

\bottomrule
\end{tabularx}
\label{tab:videophy}
\end{table*}

\textbf{Implementation Details.} Our framework requires no additional 
% learning
training of the underlying video generation models. Unless otherwise specified, DeepSeek-Pro \cite{xu2026deepseek} 
% (operated in \textit{thinking} mode)
is used for reasoning and verification. The maximum number of output tokens is set to 4,096. In the PECR module, 3 candidate successor scene graphs are generated at each reasoning step. Up to 5 events are inferred for each physical phenomenon. When predicted physical quantities violate the governing physical constraints, the reasoning is repeated for up to 3 iterations. In the TRKC module, Qwen-Image-Edit \cite{wu2025qwen} is adopted for progressive keyframe generation, using 40 sampling steps. SAM \cite{kirillov2023segment} 
% (prompted with a point predicted by DeepSeek-Pro)
is adopted to obtain masks of objects to be edited. The residual guidance coefficient $\beta$ increases from 0 to 1 following a cosine schedule as denoising proceeds. In the PCSG module, classifier-free guidance scale $\gamma$ is set to 6.0 for CogVideoX-5B \cite{yang2024cogvideox}, 5.0 for Wan2.1-VACE-1.3B \cite{wan2025wan} and Wan2.1-14B \cite{wan2025wan}, and 4.0 for Wan2.2-14B \cite{wan2025wan} respectively. For each video generator, we use its official inference configurations, including spatial resolution, number of frames, sampling steps, and sampling scheduler. All experiments are conducted on a single NVIDIA H100 GPU with 80 GB of memory.

\subsection{Quantitative Comparisons}
Tables~\ref{tab:phygenbench}, \ref{tab:videophy}, \ref{tab:physics}, and~\ref{tab:physics_iq} present quantitative comparisons on four benchmarks with complementary evaluation emphases, including physical plausibility across fundamental domains (PhyGenBench \cite{meng2024towards}), interactions among different material phases (VideoPhy \cite{bansal2024videophy}), adherence to both fundamental and anti-physics instructions (PhyWorldBench \cite{gu2025phyworldbench}), and physical process continuation from a conditioning image (Physics-IQ \cite{motamed2026generative}). In general, our framework achieves leading performance across all four benchmarks, with detailed comparisons presented below.

\begin{table}[htbp]
\centering
\small

\caption{Performance comparison on PhyWorldBench \cite{gu2025phyworldbench} across two physics
types. The best and second-best results are \textbf{highlighted} and \underline{underlined}, respectively.}
\label{tab:physics}

\resizebox{\columnwidth}{!}{%
\begin{tabular}{lcccccc}
\toprule
Methods & \multicolumn{3}{c}{Fundamental Physics (\%)} & \multicolumn{3}{c}{Anti-Physics (\%)} \\
\cmidrule(lr){2-4} \cmidrule(lr){5-7}
 & SA & PC & SA, PC  & SA & PC & SA, PC
\\
\midrule
\midrule

\multicolumn{7}{l}{\textit{Video Foundation Model}} \\
Sora-Turbo \cite{openai2024sora} & 43.8 & 31.5 & 24.6 & 13.6 & 7.8 & 3.9 \\
Gen-3 \cite{runway2024gen3alpha}  & 32.0 & 22.8 & 16.1 & 10.8 & 2.9 & 2.0 \\
Kling \cite{kling2024}  & 41.7 & 29.9 & 23.5 & 12.5 & 7.3 & 4.2 \\
Pika \cite{Pika2024} & 58.7 & \underline{37.5} & 31.2 & \underline{22.8} & 4.3 & 1.1 \\
Dream Machine \cite{LumaDreamMachine2024} & 46.4 & 25.9 & 22.0 & 5.6 & 0.0 & 0.0 \\
Hunyuan \cite{kong2024hunyuanvideo}     & 38.5 & 27.8 & 20.5 & 8.2 & 2.1 & 1.0 \\
Open-Sora-Plan \cite{zheng2024open} & 17.5 & 13.6 & 9.6 & 6.9 & 5.9 & 3.9 \\
% CogVideoX-1.5 \cite{yang2024cogvideox}     & 41.9 & 26.6 & 19.2 & 6.2 & 2.1 & 0.0 \\
Step-video-T2V \cite{ma2025step}      & 35.3 & 24.5 & 18.9 & 8.5 & 2.1 & 2.0 \\
% Wanx-2.1 \cite{wan2025wan}              & 38.9 & 27.1 & 22.0 & 8.2 & 1.7 & 1.1 \\
LTX-Video \cite{hacohen2024ltx}      & 22.1 & 10.2 & 8.0 & 7.6 & 1.1 & 0.0 \\
\midrule
\midrule
\multicolumn{7}{l}{\textit{Physics-aware Video Generation Model}} \\
CogVideoX-5B \cite{yang2024cogvideox}     & 41.9 & 26.6 & 19.2 & 6.2 & 2.1 & 0.0 \\
\rowcolor{DarkGray}
\, + Ours & \underline{61.7} & \underline{37.5} & \underline{31.4} & 21.0 & \underline{8.6} & \underline{5.7} \\
\midrule
Wan2.1-1.3B-VACE \cite{wan2025wan} & 48.9 & 27.9 & 26.4 & 18.1 & 7.6 & 3.8 \\
\rowcolor{DarkGray}
\, + Ours & 58.7 & 31.8 & 29.2 & 19.1 & 6.7 & 4.8 \\
\midrule
Wan2.2-14B \cite{wan2025wan} & 65.4 & 34.8 & 33.2 & 22.9 & 7.6 & 4.8 \\
\rowcolor{DarkGray}
\, + Ours & \textbf{72.5} & \textbf{38.7} & \textbf{36.8} & \textbf{29.5} & \textbf{10.5} & \textbf{7.6} \\

\bottomrule
\end{tabular}%
}
\end{table}

\begin{table}[htbp]
    \centering
    \caption{Performance comparisons on Physics-IQ \cite{motamed2026generative} for image conditioned generation. The best and second-best results are \textbf{highlighted} and \underline{underlined}, respectively. $^{\dagger}$ denotes reward models used for Best-of-$N$ selection among candidates generated by Wan2.2; results are reported by \cite{yuan2026inference}.}
    \label{tab:physics_iq}
    \scriptsize
    \resizebox{\columnwidth}{!}{%
    \begin{tabular}{lccccc}
        \toprule
        Methods
        & \makecell{Spatial\\\ IoU $\uparrow$}
        & \makecell{Spatio\\\ Temporal IoU $\uparrow$}
        & \makecell{Weighted \\\ Spatial IoU $\uparrow$}
        & MSE $\downarrow$
        & \makecell{Physics-IQ\\\ Score (\%) $\uparrow$} \\
        \midrule
        \midrule

        \multicolumn{6}{l}{\textit{Video Foundation Model}} \\

        Sora-Turbo \cite{openai2024sora}
        & 0.138
        & 0.047
        & 0.063
        & 0.030
        & 10.0 \\

        Pika \cite{Pika2024}
        & 0.140
        & 0.041
        & 0.078
        & 0.014
        & 13.0 \\ 

        SVD-T2I2V \cite{blattmann2023stable}
        & 0.132
        & 0.076
        & 0.073
        & 0.021
        & 14.8 \\

        VideoPoet \cite{kondratyuk2023videopoet}
        & 0.141
        & 0.126
        & 0.087
        & 0.012
        & 20.3 \\

        Lumiere \cite{bar2024lumiere}
        & 0.113
        & 0.173
        & 0.061
        & 0.016
        & 19.0 \\

        Gen-3 \cite{runway2024gen3alpha}
        & 0.201
        & 0.115
        & 0.116
        & 0.015
        & 22.8 \\

        Kling \cite{kling2024}
        & 0.197
        & 0.086
        & 0.144
        & 0.025
        & 23.6
        \\ 

        CogVideoX-1.5 \cite{yang2024cogvideox}
        & 0.198
        & \textbf{0.189}
        & 0.127
        & 0.015
        & 27.9 \\

        Wan2.2-TI2V-5B \cite{wan2025wan}
        & 0.164
        & 0.132
        & 0.102
        & 0.010
        & 22.1 \\

        VLIPP \cite{yang2025vlipp}
        & -
        & -
        & -
        & -
        & 34.6 \\

        \midrule
        \midrule
        \multicolumn{6}{l}{\textit{Physics-aware Video Generation Model}} \\

        RDPO \cite{qian2025rdpo}
        & -- & -- & -- & --
        & 25.2 \\

        PHANTOM \cite{shen2026phantom}
        & 0.245
        & 0.146
        & 0.140
        & 0.009
        & 29.6 \\

        Wan2.2-14B \cite{wan2025wan}
        & 0.358
        & 0.120
        & 0.217
        & \underline{0.008}
        & 38.3 \\

        \, + VideoMAE$^{\dagger}$ \cite{yuan2026inference}
        & 0.347
        & 0.124
        & 0.216
        & \underline{0.008}
        & 37.9 \\

         \, + Qwen2.5-VL$^{\dagger}$ \cite{yuan2026inference}
         & 0.356
         & 0.116
         & 0.213
         & 0.009
         & 37.6 \\

         \, + Qwen3-VL$^{\dagger}$ \cite{yuan2026inference}
         & 0.353
         & 0.128
         & 0.216
         & \underline{0.008}
         & 38.5 \\

         \, + WMReward$^{\dagger}$ \cite{yuan2026inference}
         & \underline{0.378}
         & \underline{0.171}
         & \underline{0.247}
         & \underline{0.008}
         & \underline{44.4} \\

         \rowcolor{DarkGray}
        \, + Ours
        & \textbf{0.405}
        & 0.160
        & \textbf{0.268}
        & \textbf{0.007}
        & \textbf{44.9}
        \\
        \bottomrule
    \end{tabular}%
    }
\end{table}

\textbf{Performance Comparisons on PhyGenBench.} As shown in Table~\ref{tab:phygenbench}, with CogVideoX-5B \cite{yang2024cogvideox}, our framework achieves PCA scores of 70.0\%, 78.7\%, 76.7\%, and 64.2\% in mechanics, optics, thermal, and material, respectively, yielding the highest average score of 72.5\%. 
% and outperforming PhysHPO \cite{chen2025hierarchical} by 11.5 percentage points. 
Applying our framework to CogVideoX-5B \cite{yang2024cogvideox}, Wan2.1-1.3B-VACE \cite{wan2025wan}, Wan2.1-14B \cite{wan2025wan}, and Wan2.2-14B \cite{wan2025wan} raises their average scores by 27.5, 10.0, 30.8, and 18.1, respectively. These gains demonstrate that our proposed event centric conditioning enhances physical plausibility in generated videos across diverse phenomena and video generation backbones.

\textbf{Performance Comparisons on VideoPhy.} As shown in Table~\ref{tab:videophy}, our approach improves physical interaction modeling across various evaluated categories. With CogVideoX-5B \cite{yang2024cogvideox}, our approach achieves SA-PC scores of 50.0\%, 63.0\%, and 63.8\% for solid-solid, solid-fluid, and fluid-fluid interactions, respectively, yielding an overall score of 58.5\% and 
% surpassing PhysHPO \cite{chen2025hierarchical} by 12.6 percentage points
surpassing the overall score of 45.9\% achieved by PhysHPO \cite{chen2025hierarchical}.
% For Wan2.2-14B \cite{wan2025wan}, the overall score increases by 18.1 percentage points, from 39.2\% to 57.3\%.
When applied to Wan2.2-14B \cite{wan2025wan}, our approach increases the overall score from 39.2\% to 57.3\%.
These improvements may partly benefit from assigning physical quantities to individual objects and their relationships across successive events. The associations clearly characterize how the states of interacting objects evolve throughout an interaction.

\textbf{Performance Comparisons on PhyWorldBench.} As shown in Table~\ref{tab:physics}, our approach consistently improves the SA-PC score across all evaluated video generators under both fundamental and anti-physics scenarios. The most substantial improvements are obtained with CogVideoX-5B \cite{yang2024cogvideox}, increasing the SA-PC scores from 19.2\% to 31.4\% for fundamental physics and from 0.0\% to 5.7\% for anti-physics. When applied to Wan2.2-14B \cite{wan2025wan}, our approach achieves the best performance in both categories, with scores of 36.8\% and 7.6\%, respectively. A higher anti-physics score indicates better adherence to unphysical instructions rather than more unintended physical violations. These gains arise from inferring semantic scene graphs directly from prompts, preserving the evolution specified by the user. Consequently, our framework can faithfully realize the user-specified dynamics in generated videos, even when they deliberately depart from real-world physics.    

\textbf{Performance Comparisons on Physics-IQ.} As shown in Table~\ref{tab:physics_iq}, under the physical continuation with image conditioning in Physics-IQ \cite{motamed2026generative}, our approach consistently improves the Wan2.2-14B \cite{wan2025wan} baseline across all metrics, increasing the overall score from 38.3\% to 44.9\%. Compared with WMReward \cite{yuan2026inference}, our approach achieves higher Spatial IoU (0.405 versus 0.378) and Weighted Spatial IoU (0.268 versus 0.247), together with a lower MSE (0.007 versus 0.008), yielding a higher overall score of 44.9\% versus 44.4\%. These gains provide evidence that visual guidance derived from event boundaries better constrains where future physical changes occur and how widely they spread.

\subsection{Qualitative Comparison}
To qualitatively assess our framework, we conduct visual analyses from several complementary perspectives, including fundamental physical phenomena generation, comparison with open source PPVG approaches, and long-chain physical phenomena generation. Across all qualitative experiments, our framework consistently adopts Wan2.2-14B \cite{wan2025wan} as the underlying video generator.

\begin{figure*}[htbp]
    \centering
    \includegraphics[width=\textwidth]{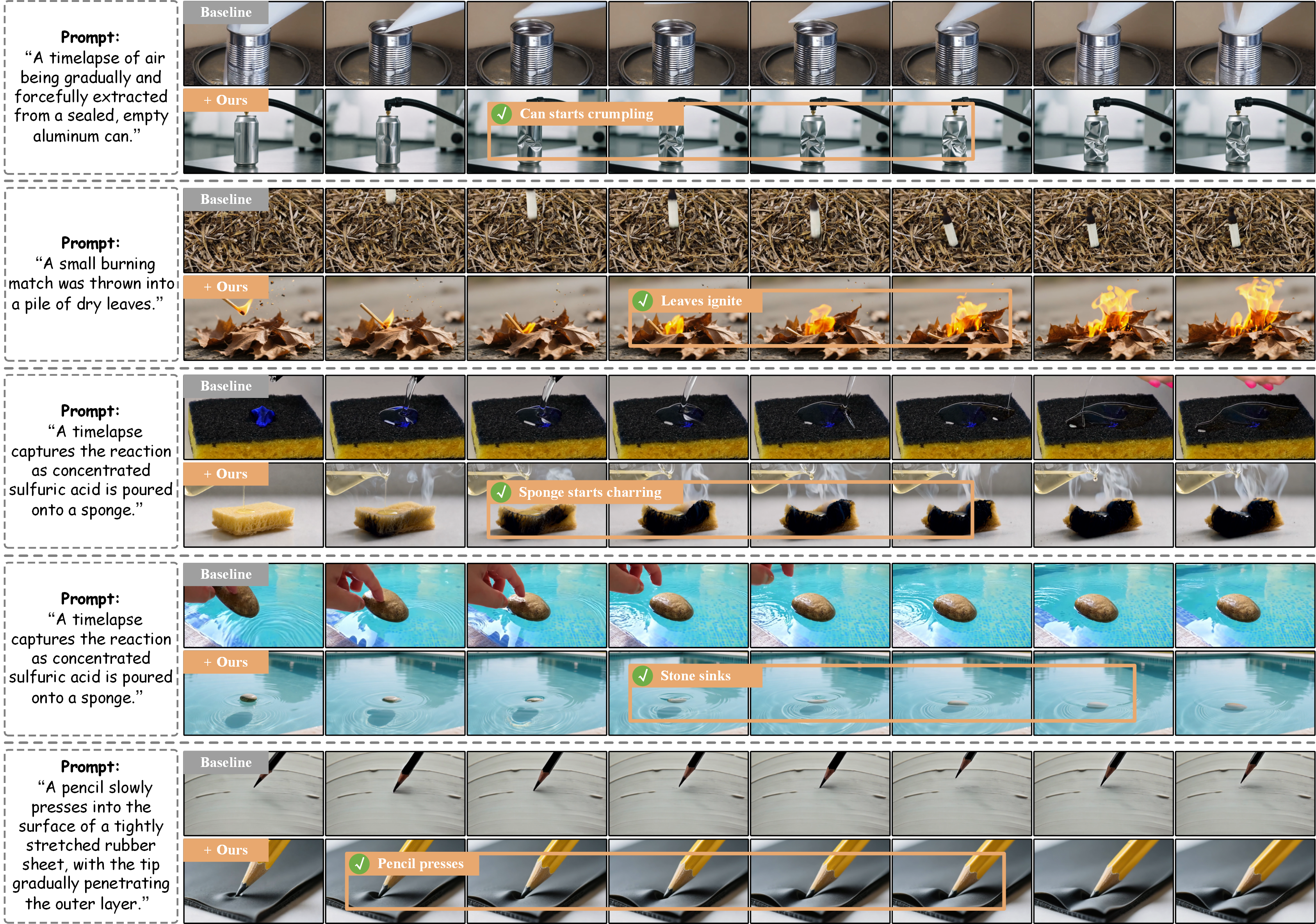}
    \caption{Visualization of generated videos across representative physical phenomena. Compared with baseline Wan2.2-14B \cite{wan2025wan}, our approach yields causally coherent progressions of physical phenomena, \textit{e.g.}, aluminum can is gradually crumpled, dry leaves ignite, sponge chars, stone sinks, and a pencil progressively presses into stretched surface. Orange boxes highlight the core physical events captured by our approach.}
    \label{fig:case_1}
\end{figure*}
\textbf{Representative Physical Phenomena Generation.} Figure~\ref{fig:case_1} presents visual comparisons across five canonical physical phenomena. Compared with baseline model Wan2.2-14B \cite{wan2025wan}, our approach more realistically reproduces progressive crumpling of an aluminum can (row 1), combustion of dry leaves (row 2), acid-induced charring of a sponge (row 3), sinking of a stone (row 4), and local deformation of a stretched membrane under pencil pressure (row 5). The benefit of our approach across such examples is primarily reflected in the completeness of physical evolution. Specifically, the baseline model typically demonstrates the delayed onset of events or skipped intermediate states. Our approach maintains a continuous progression from the triggering interaction to the induced physical response.

\begin{figure*}[htbp]
    \centering
    \includegraphics[width=\textwidth]{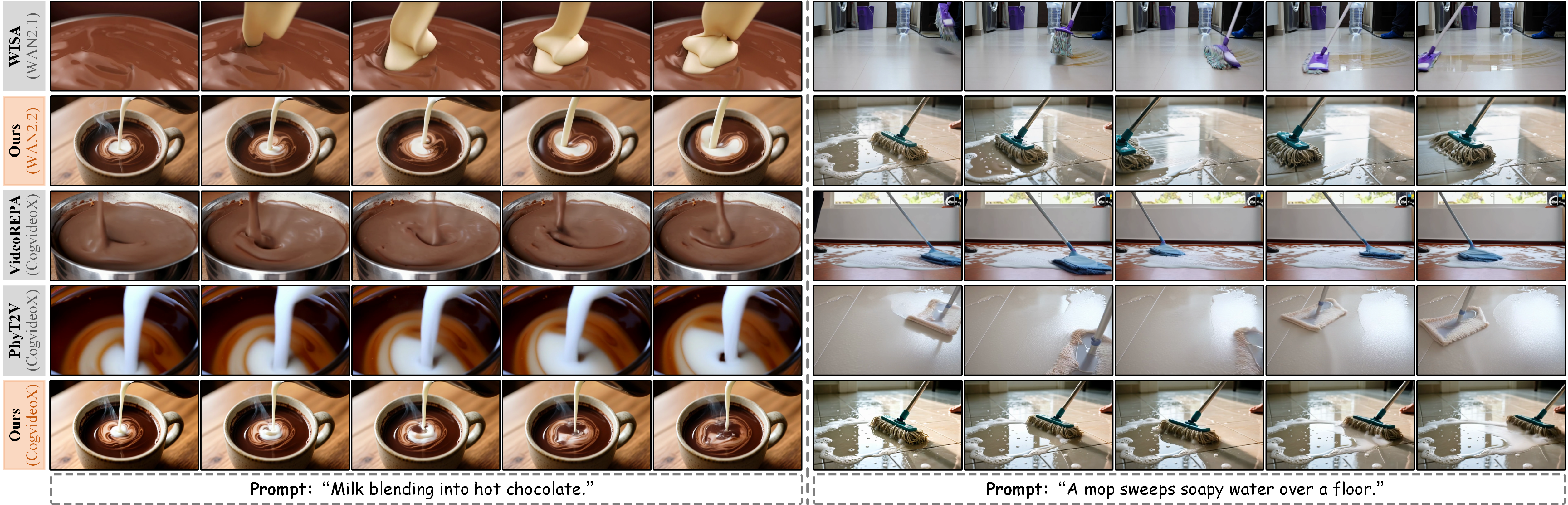}
    \caption{Visual comparison between our approach and several open source PPVG methods, primarily including WISA \cite{wang2025wisa}, VideoREPA \cite{zhang2025videorepa}, and PhyT2V \cite{xue2025phyt2v}. Comparative methods often suffer from incomplete blending or weak correspondence between mop motion and water displacement. Conversely, our approach produces more coherent physical evolution driven by physical interactions. }
    \label{fig:case_2}
\end{figure*}

\textbf{Comparison with Open-source PPVG Approaches.} Figure~\ref{fig:case_2} presents visual comparisons between our approach and open-source PPVG ones in modeling physical evolution under object interactions, with material mixing and fluid manipulation as examples. To maintain consistency with evaluation protocol shared across compared approaches, examples are drawn from the VideoPhy benchmark \cite{bansal2024videophy}. For milk mixing, current approaches often fail to depict gradual blending of milk into hot chocolate, resulting in persistent visual separation or abrupt blending. During mopping, current approaches (especially VideoREPA \cite{zhang2025videorepa} and PhyT2V \cite{xue2025phyt2v}) reproduce the sweeping motion of mop yet leave the soapy water unchanged. By comparison, our approach can produce gradual blending and visible water displacement, indicating a stronger ability to characterize physical responses induced by object interactions. 

\begin{figure}[htbp]
    \centering
    \includegraphics[width=0.48\textwidth]{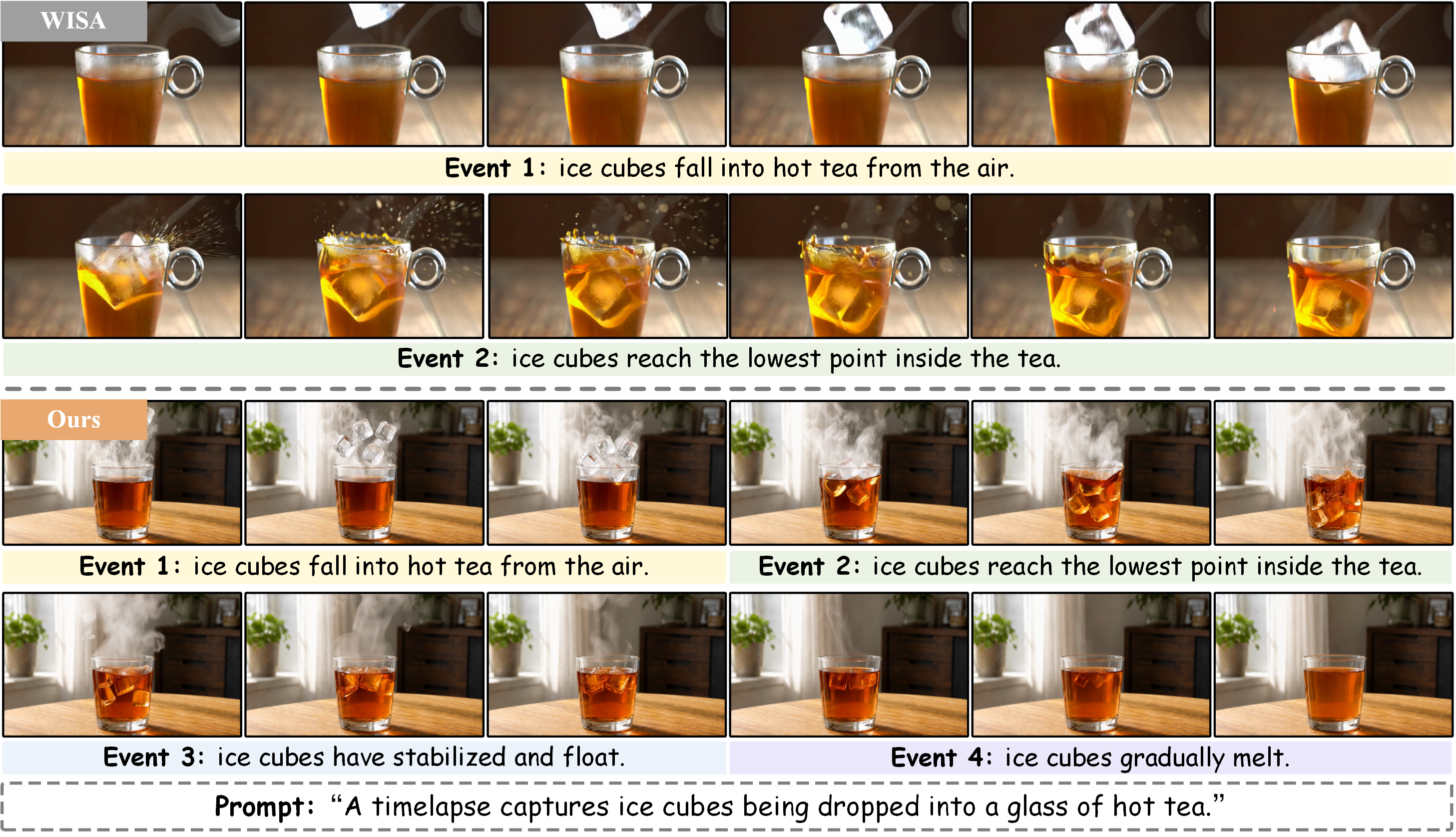}
    \caption{Visual comparison of long-chain physical phenomenon generation. Given a same prompt, WISA \cite{wang2025wisa} depicts only the early events of a phenomenon. Whereas our approach can render the complete progression from falling and sinking of ice cubes to their floating and melting. Annotations of events are added solely for visualization and are not used as model inputs. }
    \label{fig:case_3}
\end{figure}

\textbf{Long-chain Physical Phenomena Generation.} Figure~\ref{fig:case_3} presents a comparison between our approach and WISA \cite{wang2025wisa} 
% (leading PPVG approach) 
% in the generation of physical phenomena involving multiple successive events over the course of the generated video. 
in generating multiple successive physical events.
% When conditioned on the same prompt, 
Our approach can progressively model ice cubes falling into hot tea, reaching the lowest point in the liquid, stabilizing and floating, and gradually melting. Yet, WISA reproduces the early falling and splashing stages, without rendering the subsequent floating and melting events. This comparison highlights that the absence of decomposition of events causes the generative model to collapse a physical phenomenon into its most visually salient interaction, leaving other events underrepresented or omitted.

% \subsection{Ablation Studies}
\subsection{Ablation Study}
% We conduct comprehensive ablation studies to examine the effects of the proposed modules and their key design choices. Specifically, we analyze physics-driven reasoning in PECR module, specialized visual and semantic conditioning in TRKC and PCSG modules, respectively, and the sensitivity of our framework to diverse base models.
We perform comprehensive ablation studies to assess the effects of the proposed modules and their key design choices. All studies are conducted on PhyGenBench \cite{meng2024towards}, providing diverse and systematically designed physical scenarios. Besides, the PCA score considers the presence of key phenomena,  the causal order of physical events, and overall naturalness. This supports a diagnostic analysis of individual components. 

\begin{figure}
\centering
\begin{minipage}{\columnwidth}
\centering    

\captionof{table}{Ablation analysis of TRKC 
% (Section~\ref{sec:3.3}) 
and PCSG 
% (Section~\ref{sec:3.4}) 
modules.}
\label{tab:ablation_main}
\vspace{-0.4em}
\resizebox{\columnwidth}{!}{%
\begin{tabular}{ccc|cccc|c}
\toprule
\multirow{2}{*}{PECR} &
\multirow{2}{*}{TRKC} &
\multirow{2}{*}{PCSG} &
\multicolumn{4}{c|}{Physical domains (\%)} &
\multirow{2}{*}{Avg. (\%)} \\
\cmidrule(lr){4-7}
& & & 
Mechanics &
Optics &
Thermal &
Material &
\\
\midrule
\checkmark & \checkmark & $\times$   & \textbf{70.0} & 73.3 & 70.0 & 62.5 & 69.2 \\
\checkmark & $\times$   & \checkmark & 63.3 & 68.7 & 75.6 & 61.7 & 66.9 \\
\checkmark & \checkmark & \checkmark & \textbf{70.0} & \textbf{78.7} & \textbf{76.7} & \textbf{64.2} & \textbf{72.5} \\
\bottomrule
\end{tabular}%
}
\vspace{0.8em}

\includegraphics[width=\columnwidth]{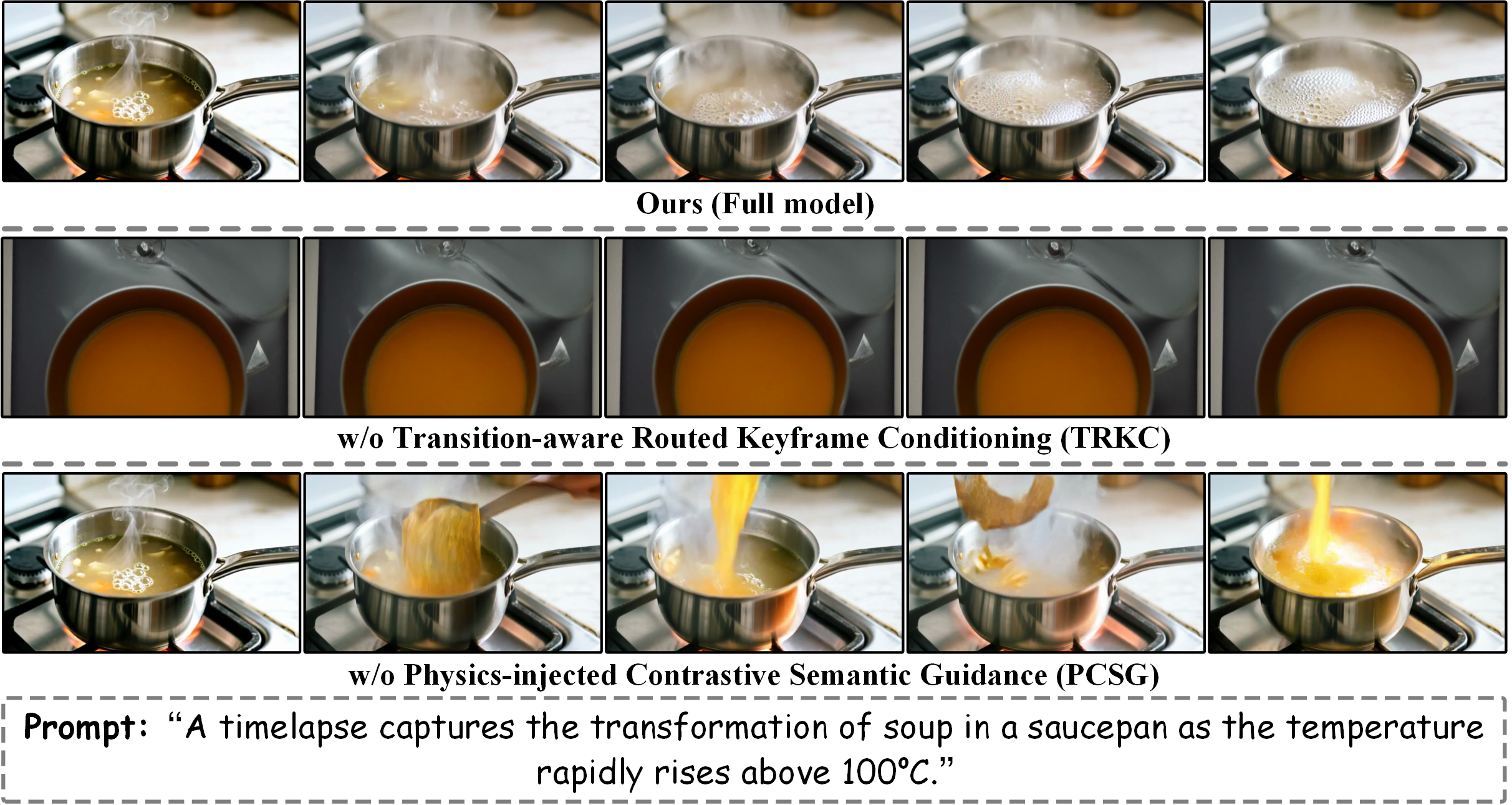}

\vspace{-0.5em}
\captionof{figure}{Qualitative ablation of TRKC
% (Section~\ref{sec:3.3}) 
and PCSG 
% (Section~\ref{sec:3.4}) 
modules. An absence of TRKC module disrupts progressive physical evolution, whereas ablating PCSG module produces implausible changes inconsistent with the user-provided description.}
\label{fig:ablation_1}
\end{minipage}
\end{figure}
\textbf{Effect of TRKC and PCSG Modules.} 
% Table~\ref{tab:ablation_main} reports an ablation analysis of individual effects of our proposed TRKC and PCSG modules. 
\indent As shown in Table~\ref{tab:ablation_main}, removing TRKC or PCSG decreases the average score from 72.5\% to 66.9\% and 69.2\%, respectively. 
This result highlights the complementary roles of TRKC module in grounding intermediate physical states and PCSG module in discriminating plausible dynamics. Figure~\ref{fig:ablation_1} also reveals the failure modes caused by ablating each module. Without TRKC module, soup remains nearly unchanged across frames, failing to depict the progressively intensifying bubbles and steam as temperature rises. This indicates the utility of TRKC module in connecting successive event states through progressive visual evolution. Without PCSG module, an unrelated pouring action followed by flames appears in the video, deviating from the specified heating process. This demonstrates the importance of PCSG module in suppressing counterfactual dynamics.

\begin{figure}[t]
\centering
\begin{minipage}{\columnwidth}
\centering

% ==================== Table ====================
\captionof{table}{
Progressive ablation analysis of PECR module. Each successive row denotes a PECR variant defined by the cumulative removal of the indicated components.
}
\label{tab:pecr_ablation}
\vspace{-0.4em}
\resizebox{\columnwidth}{!}{%
\begin{tabular}{lccccc}
\toprule
\multirow{2}{*}{Variants}
& \multicolumn{4}{c}{Physical Domains (\%)}
& \multirow{2}{*}{Avg. (\%)} \\
\cmidrule(lr){2-5}
& Mechanics
& Optics
& Thermal
& Material
& \\
\midrule
%PECR (CVPR) \cite{wang2026chain}
%& 68.3 & 75.3  & 68.9 & 61.7  & 69.0 \\
%\midrule
%\midrule
Full PECR 
& 70.0 & \textbf{78.7} & \textbf{76.7} & \textbf{64.2} & \textbf{72.5} \\
- Consistency Check
& 69.2
& \textbf{78.7}
& 75.6
& 63.3
& 71.9 \\
- Formula Reasoning 
& \textbf{71.7}
& 78.0
& 74.4
& 61.7
& 71.7 \\
- Physical Quantities
& 69.2 & 78.0 & 73.3 & 62.5 & 71.0  \\
\bottomrule
\end{tabular}%
}
\vspace{0.8em}

% ==================== Figure ====================
\includegraphics[
    width=\columnwidth
]{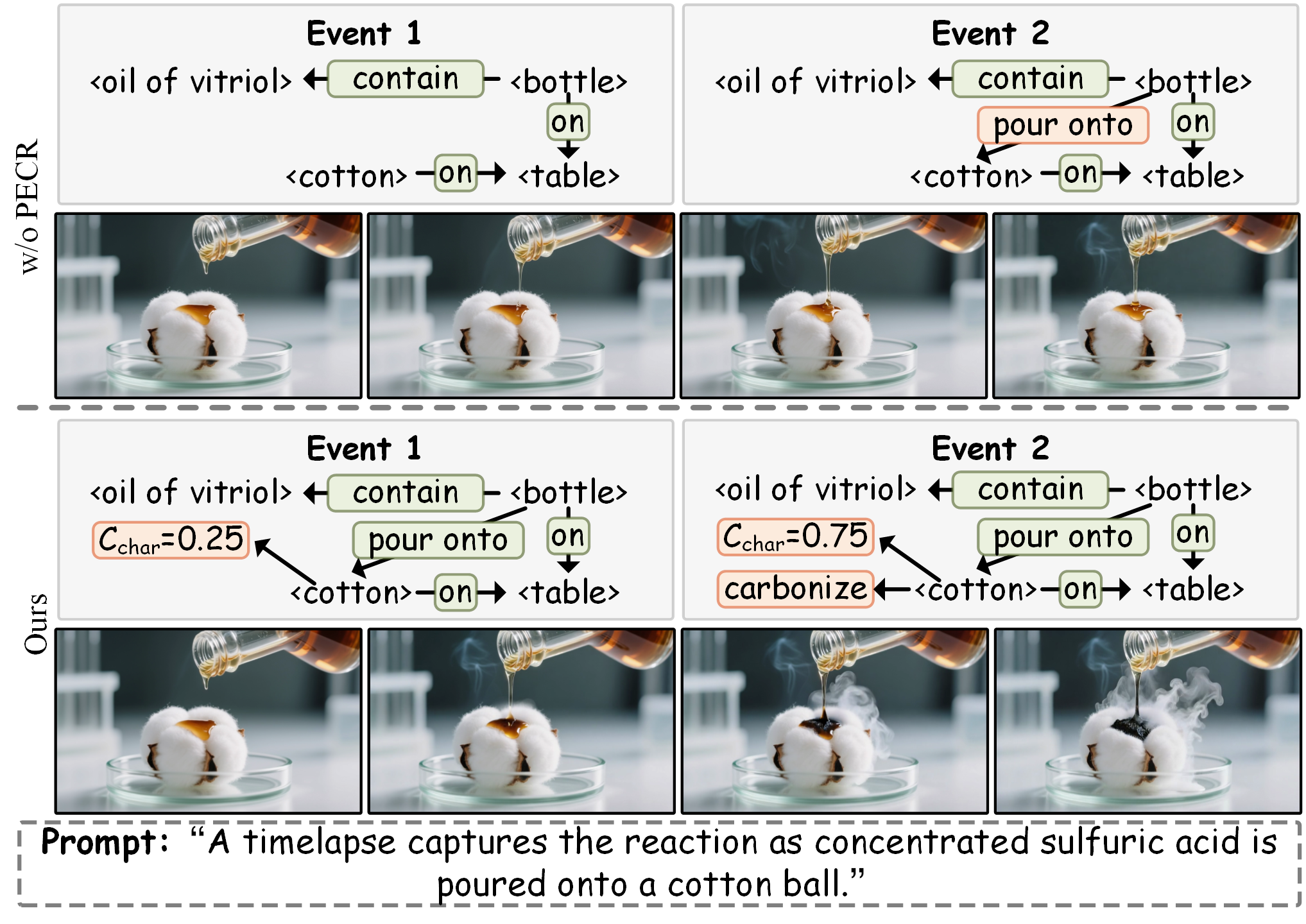}

\vspace{-0.5em}
\captionof{figure}{Visual analysis of PECR module. Physics-driven reasoning in PECR module based on applicable formulas is crucial for more accurately modeling gradual changes and preserving causal dependencies across successive events. Without PECR module, the generator becomes less responsive to variations in physical conditions.}
\label{fig:ablation_3}
\end{minipage}
\end{figure}
\textbf{Effect of Physics-driven Reasoning in PECR Module.} 
% Table~\ref{tab:pecr_ablation} summarizes the results of the progressive ablation of physics-driven reasoning in the PECR module. 
% Compared with the conference version \cite{wang2026chain}, the reformulated PECR module improves the average score from 69.0\% to 72.5\%. This improvement validates the advantage of clearly grounding physical quantities in individual objects and their relations, instead of representing physical quantities primarily as conditions associated with entire events. In addition, 
As shown in Table~\ref{tab:pecr_ablation}, the average score decreases from 72.5\% to 71.9\%, 71.7\%, and 71.0\% as consistency checking (disabling self-validation across neighboring scene graphs), formula reasoning (directly estimating numerical values of quantities by VLM), and physical quantity grounding (removing quantity from the scene graph sequence) are successively removed. 
% This consistent decline highlights the critical role of physical quantities in accurately modeling scene evolution across successive events.
This consistent decline highlights the importance of physics-driven reasoning for accurately modeling scene evolution across successive events in generated videos, with explicit physical quantity grounding providing a clear representation of object attributes and relations.

To validate the role of the PECR module in our overall framework, we disable its physics-driven reasoning and directly infer event chains from the original descriptions without explicitly modeling physical quantities. As shown in Figure~\ref{fig:ablation_3}, the reformulated PECR produces distinct material evolution under different sulfuric acid concentrations. In contrast, the variant produces only minor differences in material changes. Inspection of the inferred event chains shows that the reformulated PECR associates acid concentration with the carbonization degree of the cotton ball across successive events, whereas the variant fails to reflect this dependency. This indicates that PECR enables the evolution of events to reflect variations in physical conditions.

\begin{figure}[t]
\centering
\begin{minipage}{\columnwidth}
\centering

\captionof{table}{Ablation analysis of each 
% individual physical simulator
specialized keyframe editing operator
in TRKC module.}
\label{tab:physical_simulator}
\vspace{-0.4em}
\small
\resizebox{\columnwidth}{!}{%
\begin{tabular}{l|cccc|c}
\toprule
\multirow{2}{*}{Variants}
& \multicolumn{4}{c|}{Physical domains (\%)}
& \multirow{2}{*}{Avg. (\%)} \\
\cmidrule(lr){2-5}
& Mechanics
& Optics
& Thermal
& Material
& \\
\midrule
Baseline TRKC
& 67.5 & 76.0 & 72.2 & 58.3 & 68.8  \\
\midrule
\midrule
w Color Change
& 71.7
& 77.3
& 73.3
& 59.2
& 70.6
\\
w Texture Variation
& 70.8
& 77.3
& 74.4
& 60.8
& 71.0
\\
w Displacement
& 68.3
& 78.7
& 74.4
& 63.3
& 71.5
\\
w Deformation
& 69.2
& 78.0
& 73.3
& 63.3
& 71.3
\\
Full TRKC
& \textbf{70.0} & \textbf{78.7} & \textbf{76.7} & \textbf{64.2} & \textbf{72.5} \\
\bottomrule
\end{tabular}%
}
\vspace{0.8em}

\includegraphics[width=\columnwidth]{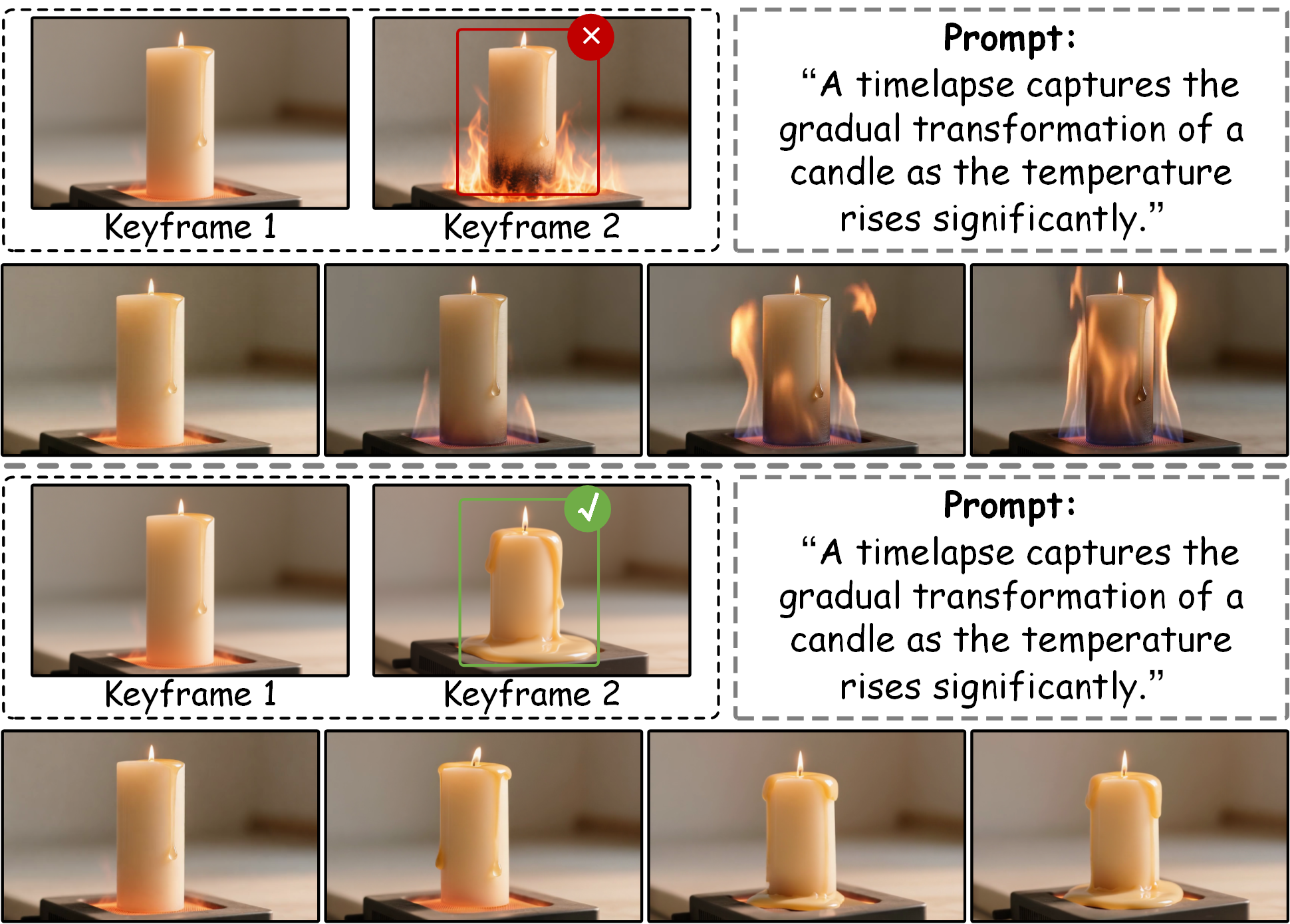}

\vspace{-0.5em}
\captionof{figure}{Visual analysis of TRKC module. Physically consistent keyframes provide reliable guidance for progressive and coherent video generation, as unrealistic visual prior misguide the generation process and lead to structurally implausible content.}
\label{fig:ablation_2}
\end{minipage}
\end{figure}
\textbf{Effect of 
% Physical Simulators
Specialized Keyframe Editing Operator
in TRKC Module.} Table~\ref{tab:physical_simulator} reports the ablation results from individually incorporating each 
% physical simulator
specialized keyframe editing operator
into baseline TRKC module (equipped only with a generic editing operator). Each specialized simulator improves upon this, with displacement simulator providing the largest gain and increasing the average score from 68.8\% to 71.5\%.These consistent improvements demonstrate the benefit of tailoring keyframe editing to the physical transition induced by each event. Full TRKC module achieves the best average score of 72.5\%, indicating the complementary roles of its specialized simulators in modeling diverse appearance changes, spatial displacements, and geometric deformations in physical phenomena. As shown in Figure~\ref{fig:ablation_2}, a deliberately erroneous keyframe causes the spurious fracture of the candle to propagate into subsequent frames, resulting in physically implausible evolution. By comparison, valid intermediate state preserves the structural integrity of the candle and reliably guides its gradual melting. This comparison highlights the importance of physically consistent keyframes for reliable video generation.

\textbf{Effect of SKGD Component in TRKC Module.} Table~\ref{tab:trkc_skgd_ablation} reports the results of ablating \textit{Soft Keyframe Guidance in Denoising} (\textit{SKGD}) from the TRKC module. Without SKGD, the average score decreases from 72.5\% to 66.3\%.
% , an overall loss of 6.2 percentage points. 
These results validate the advantage of the SKGD component over the visual prior injection scheme adopted in \cite{wang2026chain}. Unlike that scheme initializing generation with interpolated keyframe features, our SKGD module 
% characterizes feature variations between successive events and injects them residually throughout the denoising process, 
uses the differences between successive keyframes as residual guidance throughout the denoising process,
with the guidance strength varying over timesteps. This allows the denoising to progressively reconcile keyframe constraints with the generative prior.
\begin{table}[htbp]
\centering
\caption{Ablation analysis of Soft Keyframe Guidance in Denoising (SKGD) in TRKC module.}
\label{tab:trkc_skgd_ablation}
\small
\resizebox{\columnwidth}{!}{%
\begin{tabular}{l|cccc|c}
\toprule
\multirow{2}{*}{Variants}
& \multicolumn{4}{c|}{Physical domains (\%)}
& \multirow{2}{*}{Avg. (\%)} \\
\cmidrule(lr){2-5}
& Mechanics
& Optics
& Thermal
& Material
& \\
\midrule
Full TRKC
& \textbf{70.0} & \textbf{78.7} & \textbf{76.7} & \textbf{64.2} & \textbf{72.5} \\
w/o SKGD \cite{wang2026chain}
& 68.3
& 72.0
& 66.7
& 56.7
& 66.3 \\

\bottomrule
\end{tabular}%
}
\end{table}

\textbf{Effect of ENC and UCC Components in PCSG Module.} Table~\ref{tab:ablation_PSCG} investigates the core components of the PCSG module, namely \textit{Event-Chain Narrative Condensation} (\textit{ENC}) and \textit{Unphysical Counterfactual Construction} (\textit{UCC}). % Combination of the ENC and UCC components achieves the highest average score of 72.5\%, yielding absolute gains of 1.9 and 2.3 percentage points over the variants using ENC and UCC alone, respectively. 
The combination of the ENC and UCC components achieves the highest average score of 72.5\%, whereas the variants with ENC or UCC alone yield average scores of 70.6\% and 70.2\%, respectively.
% This improvement demonstrates the complementary roles of ENC (specifying desired physical evolution to be preserved during generation) and UCC (characterizing implausible physical dynamics to be suppressed) components. 
This improvement demonstrates the complementary roles of the ENC and UCC components, with ENC specifying the desired physical evolution to be preserved during generation and UCC characterizing implausible physical dynamics to be suppressed.
However, the benefit of combining ENC and UCC components is less evident in the mechanics domain. This is possibly because mechanical phenomena are often dominated by direction or velocity, which can already be sufficiently characterized by either the desired evolution or its counterfactual counterpart, leaving modest complementary information when both are used.
\begin{table}[htbp]
\centering
\caption{Ablation analysis of Event-Chain Narrative Condensation (ENC) and Unphysical Counterfactual Construction (UCC) in PCSG module.}
\label{tab:ablation_PSCG}
\resizebox{\columnwidth}{!}{%
\begin{tabular}{cc|cccc|c}
\toprule
\multirow{2}{*}{ENC} &
\multirow{2}{*}{UCC} &
\multicolumn{4}{c|}{Physical domains (\%)} &
\multirow{2}{*}{Avg. (\%)} \\
\cmidrule(lr){3-6}
& &
Mechanics &
Optics &
Thermal &
Material &
\\
\midrule
\checkmark & $\times$   & 70.8 & 77.3 & 74.4 & 59.2 & 70.6 \\
$\times$   & \checkmark & \textbf{71.7} & 76.0 & 72.2 & 60.0 & 70.2 \\
\checkmark & \checkmark & 70.0 & \textbf{78.7} & \textbf{76.7} & \textbf{64.2} & \textbf{72.5} \\
\bottomrule
\end{tabular}%
}
\end{table}

% \textbf{ENC Component vs. Causal Connectives.} 
\textbf{ENC Component vs. Causal Connective Prompting.} 
% Table~\ref{tab:ablation_PSCG_ENC} evaluates the ENC component within the PCSG module by replacing it with the causal connective strategy adopted in \cite{wang2026chain}, with the UCC component held fixed.
Table~\ref{tab:ablation_PSCG_ENC} presents a comparison between the proposed ENC component and the causal connective prompting adopted in \cite{wang2026chain}, with the UCC component held fixed. Causal connective prompting links multiple event descriptions using causal connectives, combining them into a single global semantic condition shared across all video frames.
% A clear performance gap is observed between the ENC component and the causal connective strategy, with ENC achieving a 7.1 percentage point advantage on average. 
A clear performance gap is observed. Specifically, ENC achieves an average score of 72.5\%, while the causal connective strategy achieves 65.4\%.
This 
% advantage 
improvement
% may 
arises from different organizations of condensed event semantics. To elaborate, 
% the causal connective strategy merges multiple event descriptions into a global semantic condition shared across all frames,
% (shared across all frames), 
% yet narrative ordering cannot ensure accurate assignment of individual events to their corresponding time intervals. 
although causal connectives preserve event order, the resulting global condition cannot accurately align individual events with their time intervals.
Consequently, semantic cues from different events may be simultaneously activated, causing content entanglement during generation. 
% Conversely, our framework decouples the semantic characterization of physical changes (via ENC component) from event timing (via TRKC module), enabling semantic cues to specify what changes and visual anchors to determine when. 
Conversely, our framework decouples the semantic characterization of physical changes from the timing of events. In particular, ENC provides semantic cues that specify what should change, while TRKC uses visual anchors to determine when each change should occur.
\begin{table}[htbp]
\centering
\caption{Ablation analysis of semantic conditioning strategy in PCSG module by replacing ENC component with causal connectives.}
\label{tab:ablation_PSCG_ENC}
\small
\resizebox{\columnwidth}{!}{%
\begin{tabular}{l|cccc|c}
\toprule
\multirow{2}{*}{Variants}
& \multicolumn{4}{c|}{Physical domains (\%)}
& \multirow{2}{*}{Avg. (\%)} \\
\cmidrule(lr){2-5}
& Mechanics
& Optics
& Thermal
& Material
& \\
\midrule
% PCSG module
ENC + UCC (PCSG module)
& \textbf{70.0} & \textbf{78.7} & \textbf{76.7} & \textbf{64.2} & \textbf{72.5} \\
Causal Connective \cite{wang2026chain} + UCC
& 66.7
& 68.0
& 65.6
& 60.8
& 65.4 \\

\bottomrule
\end{tabular}%
}
\end{table}

\textbf{Sensitivity to Base Models.} Table~\ref{tab:foundation_models} reports the performance of our framework with various LLMs (used in PECR and PCSG modules) and image editing models (used in TRKC module). The framework achieves consistently strong performance across the evaluated configurations, demonstrating its compatibility with heterogeneous base models. When used in PECR and PCSG modules, DeepSeek-Pro \cite{xu2026deepseek} achieves an average score of 72.5\%, surpassing gpt-oss-20b \cite{agarwal2025gpt}, Llama 3.3 \cite{touvron2023llama}, and Gemma 3 \cite{team2024gemma} by 2.7, 4.4, and 6.0 percentage points, respectively. In TRKC module, Qwen-Image-Edit \cite{wu2025qwen} improves the average score over FLUX.1 Kontext \cite{labs2025flux} and InstructPix2Pix \cite{brooks2023instructpix2pix} by 4.2 and 7.7 percentage points, respectively. These performance gaps indicate that the capability of the base model remains an important determinant of the physical plausibility of generated videos across diverse physical domains.
\begin{table}[htbp]
\centering
\caption{% Comparison across diverse base models.
Comparison of VLM choices for the PECR and PCSG modules, and image editing model choices for the TRKC module.
}
\label{tab:foundation_models}

\resizebox{\columnwidth}{!}{%
\begin{tabular}{lccccc}
\toprule
\multirow{2}{*}{Base Models}
& \multicolumn{4}{c}{PCA Score (\%)}
& \multirow{2}{*}{Avg. (\%)} \\
\cmidrule(lr){2-5}
& Mechanics
& Optics
& Thermal
& Material
& \\
\midrule

\multicolumn{6}{c}{
    \textit{Based Models Used in PECR \& PCSG Modules}
} \\

Gemma 3 12B-IT \cite{team2024gemma}
& 65.8 & 72.0 & 67.8 & 59.2 & 66.5 \\

Meta-Llama-3.3-70B-Instruct \cite{touvron2023llama}
% Llama-3.3-70B-Instruct \cite{touvron2023llama}
& 66.7 & 74.0 & 71.1 & 60.0 & 68.1 \\

gpt-oss-20b \cite{agarwal2025gpt}
& 68.3 & 75.3 & 73.3 & 61.7 & 69.8 \\

%\rowcolor{blue!8}
\rowcolor{DarkGray}
DeepSeek-Pro \cite{xu2026deepseek} (Ours)
& \textbf{70.0} 
& \textbf{78.7} 
& \textbf{76.7} 
& \textbf{64.2} 
& \textbf{72.5} 
\\

\midrule
\multicolumn{6}{c}{
    \textit{Base Models Used in TRKC Module}
} \\

FLUX.1 Kontext (dev) \cite{labs2025flux}
& 67.5 & 74.7 & 68.9 & 60.8 & 68.3 \\

InstructPix2Pix \cite{brooks2023instructpix2pix}
& 63.3 & 70.7 & 64.4 & 59.2 & 64.8 \\

%\rowcolor{blue!8}
\rowcolor{DarkGray}
Qwen-Image-Edit \cite{wu2025qwen} (Ours)
& \textbf{70.0} 
& \textbf{78.7} 
& \textbf{76.7} 
& \textbf{64.2} 
& \textbf{72.5} 
\\

\bottomrule
\end{tabular}
}
\end{table}
%\section{Limitation and Failure Cases}

\section{Conclusion and Future Work}
% This paper addresses the challenge of generating physically plausible videos faithfully depicting the causal evolution of real-world physical phenomena. 
This paper reformulates physically plausible video generation by modeling the causal evolution of physical phenomena. 
% In place of modeling a complex progression as an undifferentiated whole or characterizing an evolution with coarse semantic labels, we formulate each phenomenon as a sequence of causally connected and dynamically unfolding events. 
Specifically, we decompose each phenomenon into a causally ordered sequence of events defined by transitions between consecutive scene graphs augmented with physical quantities, instead of treating it as an undifferentiated whole.
% This chain of physical events serves as a shared basis for visual and semantic guidance used to steer video diffusion. 
This chain of events serves as a shared basis for constructing visual and semantic guidance to steer video diffusion.
% Through visual conditioning, successive scene transitions are visually grounded in a manner tailored to their diverse characteristics, enabling intermediate physical states to be coherently depicted over the course of video generation. 
Transition-aware visual conditioning accommodates diverse physical changes by establishing keyframes at event boundaries and ensuring coherent evolution between successive them.
% In parallel, semantic conditioning discriminates between physically plausible dynamics and their counterparts violating physical constraints. 
In parallel, contrastive semantic conditioning guides generation toward physically plausible dynamics and away from counterfactual alternatives that violate physical constraints.
Comprehensive experiments confirm the effectiveness of our framework in generating physically plausible videos, particularly in modeling complex and evolving physical phenomena.

% Looking ahead, several directions warrant closer investigation. 
Several promising directions can be explored in future work.
(1) \textit{Longer Horizons}. 
% We plan to
Future work may
consider generating videos of physical phenomena unfolding over longer time horizons (1$\sim$5 minutes). This requires mitigating the accumulation of upstream deviations by equipping T2V models with long- and short-term memory modules \cite{henschel2025streamingt2v}. (2) \textit{Compositional Scenes}. Our future work will investigate video generation in complex compositional scenes involving multiple interacting objects. This calls for attention refocusing \cite{wang2026training} to prevent object omission. (3) \textit{3D-Aware Dynamics}. Another promising avenue lies in grounding physical evolution within 3D space to capture contact geometry and object motion under occlusion. One possible solution is to leverage simulator-in-the-loop video generation \cite{foo2026physical}.

\balance
\bibliographystyle{IEEEtran}
\bibliography{main}
\end{document}